\documentclass[11pt, a4paper, logo, twocolumn, copyright]{deepmind}

\pdftrailerid{redacted}

\makeatletter
\renewcommand\bibentry[1]{\nocite{#1}{\frenchspacing\@nameuse{BR@r@#1\@extra@b@citeb}}}
\makeatother

\usepackage{kantlipsum, lipsum}
\usepackage{dsfont}
\usepackage{dm-colors}
\usepackage{mathtools}
\usepackage{dsfont}
\usepackage[dvipsnames]{xcolor}
\usepackage{booktabs}
\usepackage{xfrac}
\usepackage{bbm}

\usepackage{algorithm}
\usepackage{algorithmicx}

\usepackage[most]{tcolorbox}
\usepackage{xparse}
\usepackage{lipsum}
\usepackage{changepage}
\usepackage{enumitem}

\usepackage{bigdelim}
\usepackage{caption}
\usepackage{fancyvrb}
\usepackage{listings}
\usepackage{multirow}
\usepackage{pgfplots}
\usepackage[group-separator={,}]{siunitx}
\usepackage{subcaption}
\usepackage{tikz}
\usepackage{xspace}

\DeclareUnicodeCharacter{2212}{−}

\usetikzlibrary{backgrounds}
\usetikzlibrary{calc}
\usetikzlibrary{decorations.pathmorphing}
\usetikzlibrary{decorations.pathreplacing}
\usetikzlibrary{fadings}
\usetikzlibrary{fit}
\usetikzlibrary{matrix}
\usetikzlibrary{patterns}
\usetikzlibrary{positioning}
\usetikzlibrary{scopes}
\usetikzlibrary{shapes.arrows}
\usetikzlibrary{shapes.multipart}

\usepgfplotslibrary{groupplots,dateplot}
\pgfplotsset{compat=newest}

\newcommand{\squishlist}{
   \begin{list}{$\bullet$}
    { \setlength{\itemsep}{0pt}      \setlength{\parsep}{3pt}
      \setlength{\topsep}{3pt}       \setlength{\partopsep}{0pt}
      \setlength{\leftmargin}{1.5em} \setlength{\labelwidth}{1em}
      \setlength{\labelsep}{0.5em} } }

\newcommand{\squishlisttwo}{
   \begin{list}{$\bullet$}
    { \setlength{\itemsep}{0pt}    \setlength{\parsep}{0pt}
      \setlength{\topsep}{0pt}     \setlength{\partopsep}{0pt}
      \setlength{\leftmargin}{2em} \setlength{\labelwidth}{1.5em}
      \setlength{\labelsep}{0.5em} } }

\newcommand{\squishend}{
    \end{list}  }

\newcommand{\myvec}[1]{\boldsymbol{#1}}

\newcommand{\two}{(2)}

\newcommand{\ve}{\myvec{e}}

\newcommand{\vh}{\myvec{h}}

\newcommand{\vp}{\myvec{p}}

\newcommand{\vr}{\myvec{r}}

\newcommand{\vt}{\myvec{t}}

\newcommand{\ra}{\rightarrow}

\newcommand{\calD}{{\cal D}}

\newcommand{\calG}{{\cal G}}

\newcommand{\data}{\calD}

\DeclareMathAlphabet{\mathpzc}{OT1}{pzc}{m}{n}

\newcommand{\ie}{\textit{i}.\textit{e}.}
\newcommand{\eg}{\textit{e}.\textit{g}.}
\newcommand{\versus}{\textit{vs}.\,}
\newcommand{\pdata}{p_{\mbox{\footnotesize data}}}

\DeclareMathOperator{\softmax}{softmax}

\makeatletter
\tikzfading[name=fade left,
  left color=transparent!100,
  right color=transparent!0]

\newcommand{\FadeAfter}[2]{%
  \par\noindent\begin{tikzpicture}[baseline=(A.base)]
  \node[inner sep=0pt,inner ysep=0pt,outer sep=0pt,clip] (A) {\makebox[90pt][l]{#2}};
  \fill[white,path fading=fade left] ([xshift={#1}]A.south west) rectangle (A.north east);
  \end{tikzpicture}\par%
}

\makeatother

\lstdefinelanguage{proto}{
  morekeywords={bool, double, enum, message, oneof, option, repeated},
  morecomment=[l]{//},
  morestring=[b]"
}

\lstdefinestyle{mycodestyle}{
  commentstyle=\color{dmgray600},
  keywordstyle=\bfseries\color{dmblue400},
  stringstyle =\color{dmred500},
  basicstyle=\ttfamily\small,
  breakatwhitespace=false,         
  breaklines=true,                 
  captionpos=b,                    
  keepspaces=true,
  showspaces=false,                
  showstringspaces=false,
  showtabs=false,                  
  tabsize=2,
  columns=fullflexible,
  literate={0}{{\textcolor{dmblue500}{0}}}{1}%
           {1}{{\textcolor{dmblue500}{1}}}{1}%
           {2}{{\textcolor{dmblue500}{2}}}{1}%
           {3}{{\textcolor{dmblue500}{3}}}{1}%
           {4}{{\textcolor{dmblue500}{4}}}{1}%
           {5}{{\textcolor{dmblue500}{5}}}{1}%
           {6}{{\textcolor{dmblue500}{6}}}{1}%
           {7}{{\textcolor{dmblue500}{7}}}{1}%
           {8}{{\textcolor{dmblue500}{8}}}{1}%
           {9}{{\textcolor{dmblue500}{9}}}{1}%
           {0.}{{\textcolor{dmblue500}{0.}}}{1}%
           {-}{{\textcolor{dmblue500}{-}}}{1}%
           {true}{{\textcolor{dmblue500}{true}}}{1}%
}

\newtcbox{\tokenbox}{
  on line,
  colback=dmgray50,
  colframe=white,
  size=small,
  boxsep=0pt,
  boxrule=0pt,
  arc=1mm,
  frame hidden,
  fontupper=\strut\ttfamily}

\usepackage[authoryear, sort&compress, round]{natbib} 

\graphicspath{{figures/}}

\title{Computer-Aided Design as Language}

\correspondingauthor{ganin@google.com}

\keywords{computer-aided design, generative models, transformers, structured objects} 
\paperurl{}

\reportnumber{}

\renewcommand{\today}{2021-04-28} 

\author[1]{Yaroslav Ganin}
\author[1]{Sergey Bartunov}
\author[1]{Yujia Li}
\author[2]{Ethan Keller}
\author[1]{Stefano Saliceti}

\affil[1]{DeepMind}
\affil[2]{Onshape}

\begin{abstract}
Computer-Aided Design (CAD) applications are used in manufacturing to model everything from coffee mugs to sports cars. These programs are complex and require years of training and experience to master. A component of all CAD models particularly difficult to make are the highly structured 2D sketches that lie at the heart of every 3D construction. In this work, we propose a machine learning model capable of automatically generating such sketches. Through this, we pave the way for developing intelligent tools that would help engineers create better designs with less effort. Our method is a combination of a general-purpose language modeling technique alongside an off-the-shelf data serialization protocol. We show that our approach has enough flexibility to accommodate the complexity of the domain and performs well for both unconditional synthesis and image-to-sketch translation.
\end{abstract}

\begin{document}

\maketitle

\section{Introduction}
Computer-Aided Design (CAD) is used in the production of most everyday objects: from cars to robots to stents to power plants. CAD has replaced pencil drawings with precise computer sketches, enabling unparalleled precision, flexibility, and speed. Despite these improvements the CAD engineer must still develop, relate and annotate all the minutiae of their designs with the same attention to detail as their drafting-table forebears. The next step change in CAD productivity will come from the careful application of machine learning to automate predictable design tasks and free the engineer to focus on the bigger picture. The flexibility and power of deep learning is uniquely suited to the complexity of design.

\begin{figure}[p]
    \centering
    \vspace{-10mm}
    \input{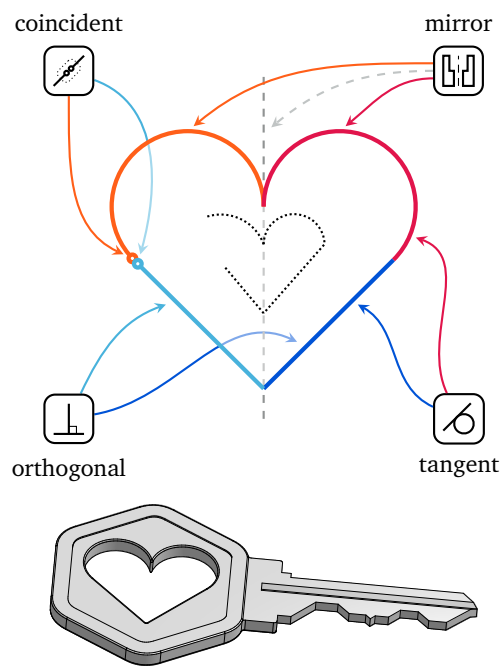}%
    \caption{\textbf{The anatomy of a CAD sketch}. Sketches are the main building block of every 3D construction. A sketch consists of geometric \emph{entities} (primitives like lines and arcs) and \emph{constraints} (\eg, tangent and mirror). The constraints are meant to convey the \emph{design intent} and define how the shape should change under various transformations of the entities. The \emph{dotted sketch} demonstrates what happens if we drop some constraints and modify the parameters of the primitives -- the design idea is lost.}
    \label{fig:teaser}
    \vspace{-10mm}
\end{figure}

Sketches are at the heart of mechanical CAD. They are the skeleton from which three dimensional forms are made. A sketch consists of various lines, arcs, splines and circles related by specific constraints such as tangency, perpendicularity and symmetry. Figure \ref{fig:teaser} illustrates how constraints relate different geometric entities to create well-defined shapes. The dotted lines show another equally valid solution to the heart if some of the constraints were dropped. The geometric entities all lie on a single sketch plane and together form enclosed regions used by subsequent construction operations such as lofts and extrusions to generate complex 3D geometry.

Constraints make sketches much more complicated than they initially appear. They express relationships that can indirectly affect every entity in the sketch. For instance, in Figure \ref{fig:teaser}, if the point at which the two arcs meet is dragged up while the bottom corner is kept stationary, the heart will grow in size. Even though the transformation may seem simple, that growth is actually the result of every constraint working in concert. The constraints ensure the shape remains in the state the designer conveyed despite every one of the entities having changed size and location. Because of this complex interplay between entities, it's easy to accidentally specify a set of constraints that is impossible for the constraint solver to satisfy, resulting in an invalid sketch. For example, two lines that satisfy both parallel and perpendicular constraints are impossible to draw. In a complex sketch, chains of constraint dependencies can make it exceedingly difficult for a designer to determine which constraint to add without invalidating the sketch. Additionally, for a given set of entities, there are many equally valid constraint systems that yield similar sketch behavior. A high quality sketch uses a set of constraints that preserves design intent, meaning the semantics of the sketch are preserved even as entity parameters (such as dimensions) are altered \citep{kyratzi20, barbero17,wu09,hartman05}. Simply put, no matter how the entity dimensions are changed, the heart in \ref{fig:teaser} will always be a heart. The challenge of capturing design intent coupled with the inherent complexity of choosing a consistent constraint system make sketch generation an exceedingly difficult problem \citep{rynne08}.

The aforementioned complexities of sketch construction are analogous to those of natural language modeling. Selecting the next constraint or entity in a sketch is like the generation of the next word in a sentence. In both contexts, the selection must function grammatically (form a consistent constraint system in the case of the sketch) and work towards some cohesive meaning (preserve design intent). Luckily, there are many tools that have proved highly successful in generating natural language. The top performers are unequivocally machine learning models that train on vast amounts of real-world data \citep{radford19,brown20}. The Transformer architecture \citep{vaswani17} in particular demonstrates unparalleled ability to create cohesive sentences, and is therefore a promising choice for adapting to the task of sketch generation. This work is our take at this adaptation. We make the following contributions:

\begin{figure*}[t!]
    \centering
    \tikzset{base style/.style={
	rounded corners, draw, thick,
	minimum width=0.3cm,
	anchor=base}}
\tikzset{timestep style/.style={base style, font=\vphantom{Ag}}}
\tikzset{state style/.style={timestep style, draw=none, fill opacity=0.25, text opacity=1.0}}	
\tikzset{output style/.style={base style, draw=none}}
\tikzset{transformer style/.style={timestep style, draw=none, minimum height=1cm}}
\tikzset{snake style/.style={-stealth, thick, decorate, decoration={snake, amplitude=.4mm, segment length=2mm, post length=2mm}}}

\tikzset{pics/brk/.style 2 args={code={
	\path [fill=#1] (-1mm,-1mm) -- (0,1mm) -- (1mm,1mm) -- (0,-1mm) -- cycle;
	\path [thick, draw, #2] (-1mm,-1mm) -- (0,1mm);
	\path [thick, draw, #2] (0,-1mm) -- (1mm,1mm);
}}}

\tikzset{tangent pic/.pic={
	\matrix (bbox) [inner sep=0] {
	\begin{scope}[rotate=45, scale=0.3]
    	\draw [thick] (0,0) circle (0.5cm);
    	\draw [thick] (-1cm,0.5cm) -- (1cm,0.5cm);
	\end{scope} \\
	};
}}

\begin{tikzpicture}

\matrix [column sep=6.4mm, row sep=5mm] {
\node (s1) [state style, fill] {}; &
\node (s2) [state style, fill=dmblue400] {}; &
\node (s3) [state style, fill=dmblue400] {}; &
\node (s4) [state style, fill=dmblue400, minimum width=3.4cm] {\texttt{line.end.x}}; &
\node (s5) [state style, fill=dmblue400, minimum width=3.4cm] {\texttt{line.end.y}}; &
\node (s6) [state style, fill=dmred500] {}; &
\node (s7) [state style, fill=dmred500] {}; &
\node (s8) [state style, fill=dmpurple500] {}; &
\node (s9) [state style, fill=dmpurple500] {}; &
\node (s10) [state style, fill] {}; &
\node (s11) [state style, fill] {}; \\

& & & \node (o4) [output style, minimum width=3cm] {(0, \textbf{0.6}, False)}; \\

\node (tr1) [transformer style] {}; &
\node (tr2) [transformer style] {}; &
\node (tr3) [transformer style] {}; &
\node (tr4) [transformer style] {}; &
\node (tr5) [transformer style] {}; &
\node (tr6) [transformer style] {}; &
\node (tr7) [transformer style] {}; &
\node (tr8) [transformer style] {}; &
\node (tr9) [transformer style] {}; &
\node (tr10) [transformer style] {}; &
\node (tr11) [transformer style] {}; \\

\node (in1) [timestep style] {}; &
\node (in2) [timestep style] {}; &
\node (in3) [timestep style, color=dmblue400] {}; &
\node (in4) [timestep style, color=dmblue400, minimum width=3.4cm] {\texttt{line.start.y:\,\scriptsize$ \ldots $}}; &
\node (in5) [timestep style, color=dmblue400, minimum width=3.4cm] {\texttt{line.end.x:}\,\footnotesize\textbf{0.6}}; &
\node (in6) [timestep style, color=dmblue400] {}; &
\node (in7) [timestep style, color=dmred500] {}; &
\node (in8) [timestep style, color=dmred500] {}; &
\node (in9) [timestep style, color=dmpurple500] {}; &
\node (in10) [timestep style, color=dmpurple500] {}; &
\node (in11) [timestep style] {}; \\
};

\draw [-stealth, thick] (in1) to (tr1);
\draw [-stealth, thick] (in2) to (tr2);
\draw [-stealth, thick, color=dmblue400] (in3) to (tr3);
\draw [-stealth, thick, color=dmblue400] (in4) to (tr4);
\draw [-stealth, thick, color=dmblue400] (in5) to (tr5);
\draw [-stealth, thick, color=dmblue400] (in6) to (tr6);
\draw [-stealth, thick, color=dmred500] (in7) to (tr7);
\draw [-stealth, thick, color=dmred500] (in8) to (tr8);
\draw [-stealth, thick, color=dmpurple500] (in9) to (tr9);
\draw [-stealth, thick, color=dmpurple500] (in10) to (tr10);
\draw [-stealth, thick] (in11) to (tr11);

\draw [-stealth, thick] (tr1) to (s1);
\draw [-stealth, thick] (tr2) to (s2);
\draw [-stealth, thick, color=dmblue400] (tr3) to pic [sloped, pos=0.5] {brk={white}{solid,-}} (s3);
\draw [-stealth, thick, color=dmblue400] (tr4) to (o4);
\draw [-stealth, thick, color=dmblue400] (o4) to (s4);
\draw [-stealth, thick, color=dmblue400] (tr5) to (s5);
\draw [-stealth, thick, color=dmblue400] (tr6) to (s6);
\draw [-stealth, thick, color=dmred500] (tr7) to pic [sloped, pos=0.5] {brk={white}{solid,-}} (s7);
\draw [-stealth, thick, color=dmred500] (tr8) to (s8);
\draw [-stealth, thick, color=dmpurple500] (tr9) to pic [sloped, pos=0.5] {brk={white}{solid,-}} (s9);
\draw [-stealth, thick, color=dmpurple500] (tr10) to (s10);
\draw [-stealth, thick] (tr11) to (s11);

\draw [snake style] (s1) to pic [pos=0.45] {brk={dmgray50}{-}} (s2);
\draw [snake style, color=dmblue400] (s2) to pic [pos=0.45] {brk={dmgray50}{-}} (s3);
\draw [snake style, color=dmblue400] (s3) to (s4);
\draw [snake style, color=dmblue400] (s4) to (s5);
\draw [snake style, color=dmblue400] (s5) to (s6);
\draw [snake style, color=dmred500] (s6) to pic [pos=0.45] {brk={dmgray50}{-}} (s7);
\draw [snake style, color=dmred500] (s7) to (s8);
\draw [snake style, color=dmpurple500] (s8) to pic [pos=0.45] {brk={dmgray50}{-}} (s9);
\draw [snake style, color=dmpurple500] (s9) to (s10);
\draw [snake style] (s10) to pic [pos=0.45] {brk={dmgray50}{-}} (s11);

\draw [-stealth, thick, dashed] (s1) to [out=-45,in=135] pic [sloped, pos=0.287] {brk={white}{solid,-}} (in2);
\draw [-stealth, thick, dashed, color=dmblue400] (s2) to [out=-45,in=135] (in3);

\node (fake_in4) at (in4.west) [timestep style, draw=none, anchor=west] {};
\node (fake_s4) at (s4.east) [timestep style, draw=none, anchor=east] {};
\node (fake_in5) at (in5.west) [timestep style, draw=none, anchor=west] {};
\node (fake_s5) at (s5.east) [timestep style, draw=none, anchor=east] {};

\node at ($(in1)!0.5!(in2)$) [inner sep=0, anchor=center] {\scriptsize$ \ldots $};
\node at ($(in3)!0.5!(fake_in4)$) [inner sep=0, anchor=center] {\scriptsize$ \color{dmblue400}\ldots $};
\node at ($(in7)!0.5!(in8)$) [inner sep=0, anchor=center] {\scriptsize$ \color{dmred500}\ldots $};
\node at ($(in9)!0.5!(in10)$) [inner sep=0, anchor=center] {\scriptsize$ \color{dmpurple500}\ldots $};
\node at ($(in10)!0.5!(in11)$) [inner sep=0, anchor=center] {\scriptsize$ \ldots $};

\draw [-stealth, thick, dashed, color=dmblue400] (s3) to [out=-45,in=135] (fake_in4);
\draw [-stealth, thick, dashed, color=dmblue400] (fake_s4) to [out=-45,in=135] (fake_in5);
\draw [-stealth, thick, dashed, color=dmblue400] (fake_s5) to [out=-45,in=135] (in6);

\draw [-stealth, thick, dashed, color=dmred500] (s6) to [out=-45,in=135] (in7);
\draw [-stealth, thick, dashed, color=dmred500] (s7) to [out=-45,in=135] (in8);
\draw [-stealth, thick, dashed, color=dmpurple500] (s8) to [out=-45,in=135] (in9);
\draw [-stealth, thick, dashed, color=dmpurple500] (s9) to [out=-45,in=135] (in10);
\draw [-stealth, thick, dashed] (s10) to [out=-45,in=135] pic [sloped, pos=0.287] {brk={white}{solid,-}} (in11);

\node (transformer) [fit=(tr1) (tr11), rounded corners, draw, thick, fill=white, fill opacity=0.75, inner sep=0pt] {};
\node at (transformer.center) {Transformer \citep{vaswani17} + Pointer Net \citep{vinyals15}};

\begin{scope}[on background layer]
	\node (states) [fit=(s1) (s11), rounded corners, fill=dmgray50, inner sep=1mm] {};
	\node (tokens) [fit=(in1) (in11), rounded corners, fill=dmgray50, inner sep=1mm] {};
\end{scope}

\node at (states.north west) [anchor=south west] {\small\textcolor{dmgray400}{Interpreter states}};
\node at (tokens.south west) [anchor=north west] {\small\textcolor{dmgray400}{Tokens}};

\begin{scope}[local bounding box=heart, shift={($(tokens.south) - (1.2cm,2.2cm)$)}, yscale=-1]
	\coordinate (heart_bottom) at (0,2.4142135623730945cm);
    \coordinate (heart_joint2) at (1.7071067811865475cm, 0.7071067811865476cm);
	\draw [fill=white, draw, fill opacity=0.5, thick] (0, 0) arc (0:-225:1cm) coordinate (heart_joint1) -- (heart_bottom) -- (heart_joint2) arc (225:0:-1cm) -- cycle;
	\draw [draw=dmred500, ultra thick] (heart_joint2) arc (225:0:-1cm) node (arc_entity) [pos=0.5, inner sep=0pt, minimum size=1mm] {} node (arc_entity2) [pos=0.15, inner sep=0pt, minimum size=1mm] {};
	\draw [draw=dmblue400, ultra thick] (heart_bottom) -- (heart_joint2) node (line_entity) [pos=0.8, inner sep=0pt, minimum size=1mm] {};
\end{scope}

\draw [thick, decoration={brace, mirror, raise=0.25cm}, decorate, color=dmblue400] (in3.south west) -- (in6.south east) node (line_tokens) [midway, anchor=north, minimum size=4mm] {};
\draw [thick, decoration={brace, mirror, raise=0.25cm}, decorate, color=dmred500] (in7.south west) -- (in8.south east) node (arc_tokens) [midway, anchor=north, minimum size=4mm] {};
\draw [thick, decoration={brace, mirror, raise=0.25cm}, decorate, color=dmpurple500] (in9.south west) -- (in10.south east) node (tangent_tokens) [midway, anchor=north, minimum size=4mm] {}; 

\begin{scope}[on background layer]
	\draw [-stealth, thick, dotted, color=dmblue400] (line_tokens) to [out=270, in=135] (line_entity);
	\draw [-stealth, thick, dotted, color=dmred500] (arc_tokens) to [out=270, in=70] (arc_entity);
\end{scope}

\pic (tangent_icon) [right=1cm of heart.east, color=dmpurple500] {tangent pic};
\node (tangent) [fit=(tangent_iconbbox), rounded corners, draw, ultra thick, inner sep=1mm, color=dmpurple500] {};

\node (tangent_label) [below=1mm of tangent.south] {\textcolor{dmpurple500}{tangent}};

\draw [-stealth, thick, color=dmblue400] (tangent) to [out=190, in=-20] (line_entity);
\draw [-stealth, thick, color=dmred500] (tangent) to [out=170, in=-10] (arc_entity2);

\draw [-stealth, thick, dotted, color=dmpurple500] (tangent_tokens) to [out=270, in=70] (tangent);

\end{tikzpicture}
    \caption{\textbf{Interpreter-guided generation of a sketch.} At each point in time, a Transformer \citep{vaswani17} outputs a raw value which is fed into an \emph{interpreter} that decides which field of a \emph{Protocol Buffers} message this value corresponds to. Once the field is populated the interpreter communicates its decision back to the Transformer and transitions to the next state. The system is capable of generating sequences of \emph{arbitrary structured objects} and therefore is suitable for synthesizing both sketch entities and constraints in one go.} 
    \label{fig:main}
\end{figure*}
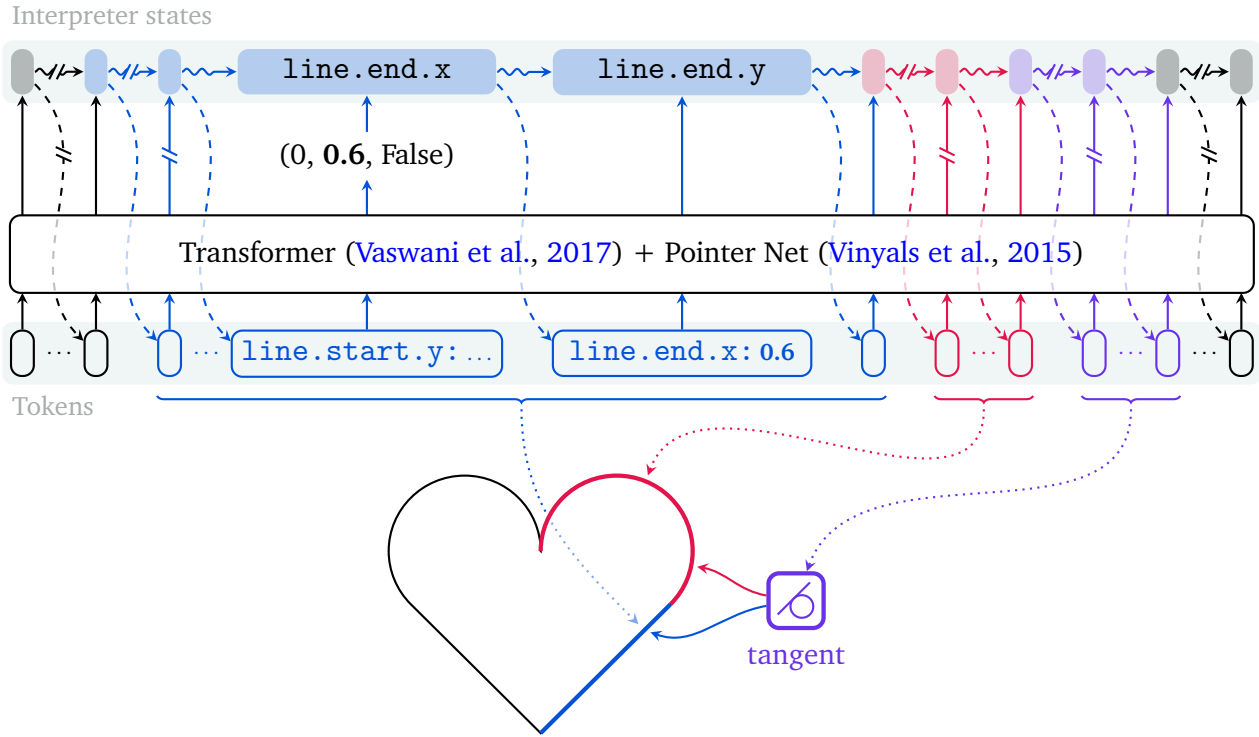

\begin{itemize}
    \item We devise a method for describing structured objects using Protocol Buffers \citep{varda08} and demonstrate its flexibility on the domain of natural CAD sketches.
    \item We propose several techniques for capturing distributions of objects represented as serialized Protocol Buffers. Our approach draws inspiration from recent advances in language modeling while focusing on eliminating data redundancy.
    \item We collect a dataset containing over 4.7M of carefully preprocessed parametric CAD sketches. We use this dataset to validate the proposed generative models. To our knowledge, the experiments presented in this work significantly surpass the scale of those reported in the literature both in terms of the amount of training data and the model capacity.
\end{itemize}

\section{Related work}
\label{sec:related_work}
\paragraph{Constraints and design intent.}

Well-chosen sketch constraints are essential to properly convey design intent~\citep{Ault99} and facilitate its resilience to successive parameters modifications which is often understood as a measure of the quality of a design document~\citep{company19}. Among studies conducted on how to predict constraints in a CAD sketch, \citet{kyratzi20} introduce an analysis of intention-regularities and the relative set of meta-constraints, thus prescribing a set of rules to properly design a constraint set. \citet{veuskens20} present a framework to iteratively guide the user in suggesting a set of missing constraints from a sketch. None of the proposed approaches, derived from the Human-Computer Interaction community, leverages machine learning techniques to infer those rule-sets from the data. Since explicit programming of such rules can be very difficult, performance of rule-based systems is constrained by the effort put in curating the set of rules which, we believe, is ultimately less efficient than directly capturing human practice by a high-capacity generative model.
 
\paragraph{Datasets and generative models for CAD.}

Until recently there were very few parametric CAD datasets large and varied enough to serve as training data for machine learning. This situation had started to change with the release of the ABC dataset~\citep{koch19}. It contains a collection of 3D shapes from the Onshape public repository \citep{onshape} and is intended to be used in geometric deep learning research. In order to enable the latter, the authors provide ground-truth targets for a variety of tasks ranging from estimation of differential surface properties to feature detection. Unfortunately, the main focus of this work revolves around meshes and, as a result, the parametric aspect of the data is largely overlooked. This manifests itself in incompleteness of the supplied symbolic representations of the shapes. When it comes to the CAD sketches specifically, little to no effort was put to extract and prepare them for machine learning applications.

Several works concurrent with ours aim to address the limitations of~\citep{koch19} and shift focus to CAD construction sequences, \ie, to \emph{how} of design rather than \emph{what}. \citet{seff20} center their attention on contributing a more complete and easy to use dataset of 2D sketches. As we discuss in Section~\ref{sec:experiments}, the resulting data suffers from a significant amount of duplication and therefore should be used with caution. In addition to the dataset, the paper also proposes a baseline generative model employing autoregressive message passing networks \citep{gilmer17}. This approach, however, can only handle simple sketches and heavily relies on the external solver to determine the final placement of primitives.

Fusion 360 Gallery \citep{willis20} attacks CAD data from a different angle. Here, the task is to recover a sequence of extrusion operations that gives rise to a particular target 3D shape. Despite dealing with 3D, this setting is deliberately limited: sketches are assumed to be given and the proposed model only decides on which sketch to extrude and to what extent. While \citep{willis20} is a step towards full parametric CAD generation, it is unclear how well this approach will scale to more real-world scenarios.

\paragraph{Image to CAD conversion.}

Beyond unconditional generation, the research community has a long-standing interest in the problem of conversion from a drawing (either computer-rendered or hand-drawn) to a vectorized CAD description.
This task can be seen as a special case of image vectorization~\citep{jimenez82} and has been extensively studied in the past~\citep{nagasamy90,vaxiviere92,dori95,dori99}. However, those systems mostly relied on heuristic object recognition  and were not learning-driven. In this paper we present an image conditional model as a solution to this problem. In contrast to existing approaches we make no assumptions on visual appearance, do not program any object recognition heuristics, and instead learn them from the training data.
 
\paragraph{Vector image generation and inference.}

Synthesizing CAD sketches bears a lot of similarities with predicting vector graphics. In this field, several recent works employ autoencoders to model vector art. \citet{carlier20} present a hierarchical VAE \citep{kingma13} capable of producing novel SVG icons. Im2Vec \citep{reddy21} takes this idea even further and rely on a differentiable renderer to remove the necessity for having training data in vector format. In contrast to our approach, both of these methods use highly domain-dependent architectures and, therefore, it would be a non-trivial task to adapt them for generation of complex sketch objects. 
Similarly,~\citep{li20} use differentiable renderers to allow gradient-based optimization of visible entities constituting a vector image. However, this idea is not straightforwardly applicable to sketch generation since the choice of what entities to use in the sketch as well as handling of constraints are discrete in their nature.

\citet{egiazarian20} address the task of vectorizing raster technical drawings by means of a custom, multi-stage computer vision pipeline.
As such, this system is not formulated as an end-to-end trainable model and also does not concern constraint inference which, as we argue, is an important aspect of sketch modeling.

\paragraph{Transformers for sequence modeling.}

In recent years, Transformers \citep{vaswani17} have become the dominating approach in many sequence modeling applications. Works like \citep{razavi19,dhariwal20,brown20,ramesh21} demonstrate impressive and sometimes surprising capability of the architecture to exhibit intelligent behaviour in tasks like image and audio synthesis and generation of natural language. For that reason, we employ Transfomers as a computational backbone for the system we present in this manuscript. 

Our method can be seen as generalization of PolyGen \citep{nash20}, a Transformer-based generative model for 3D meshes. Similarly to \citep{nash20}, we use Pointer Networks \citep{vinyals15} to relate items in the synthesized sequence. Unlike PolyGen, however, our framework can handle non-homogeneous structures of arbitrary complexity. Moreover, we simplify the architecture to use a single neural network to generate the entire object of interest. All these improvements make our approach a good fit for modeling CAD sketches and potentially other components of CAD constructions.

\section{Data}
\subsection{Sketches}

\begin{figure}[t!]
    \centering
    \includegraphics[width=0.9\linewidth]{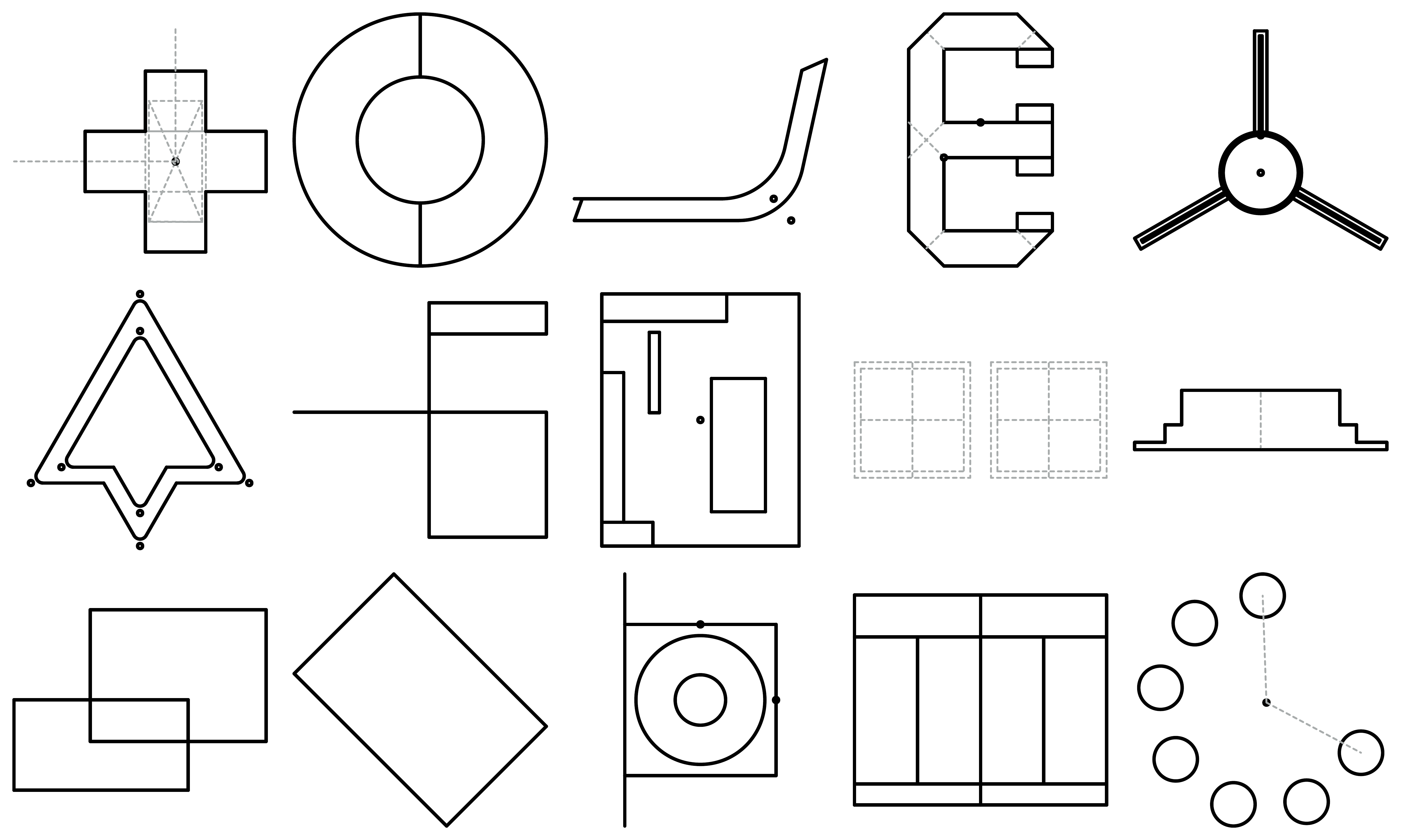}
    \caption{\textbf{Examples from our sketch dataset.} Here, we only show \emph{entities}. Although typical sketch shapes are somewhat regular, the data contains a significant number of objects that may seem invalid. The latter is usually due to the fact that we are considering them outside the context of the 3D construction they belong to.}
    \label{fig:dataset_examples}
\end{figure}

A CAD \emph{sketch} is a 2D drawing that lives on some 3D plane and consists of a set of geometric primitives such as points, line segments, arcs and circles called \emph{sketch entities} (see Figure~\ref{fig:dataset_examples}). Sketches are considered a starting point for most CAD models: a 2D sketch can be ``extruded'' to create a 3D shape; further sketches can be used to modify and refine existing 3D geometries by creating pockets, ridges, holes or protrusions. The sketching operation in Onshape accounts for about 35\% of the features created every day \citep{chastell15} making it a potentially high impact target for machine learning research.

Onshape is a \emph{parametric} solid modeling software. This means that designs are dimension-driven, \ie, the geometry can be changed by modifying dimensions. Ideally, we would like these changes to be predictable and to preserve the integrity of the geometry, \eg, connected line segments should remain connected, orthogonal curves should remain orthogonal and so on. In other words, sketch transformations should respect the \emph{design intent}. One way to achieve this is by introducing a set of relationships between sketch entities. For example, if we draw two circles that happen to be concentric we can state this relation explicitly so that when one of the circles is moved, the other moves with it, maintaining the relation. We call such relations \emph{sketch constraints}. In this framework, in order to create a simple rectangular shape we would start with four line segments, constrain the end points to coincide forcing the shape to stay closed at all times and, finally, make the opposite and the adjacent segments parallel and orthogonal respectively.

A 2D sketch is thus defined by two collections of objects: entities and constraints. The parameters of entities (\eg, coordinates of points, radii of circles and so on) that we supply to the CAD software do not need to be precise as they will be used only as an \emph{initial guess} by the underlying \emph{constraint solver}. The solver's task is to adjust the geometric configuration so that all the constraints are satisfied. It should be noted that, in general, the solver considers entities and constraints holistically so the order in which they appear in corresponding collections does not matter. As we model sketches as sequences, this does mean that there are many different orderings (and also potentially different parameter values for the entities, as they do not need to be precise) that correspond to the same sketch. We will expand more on this in Section~\ref{sec:data_layout}.

\subsection{Data layout}
\label{sec:data_layout}

In order to store and process sketches, we employ Protocol Buffers (PB) \citep{varda08} rather than the raw JSON format \citep{pezoa16} provided by the Onshape API. The benefit of using this approach is twofold: the resulting data occupies less space because unnecessary information is removed, and the PB language makes it easy to define precise specifications for complex objects with varied structure.

\begin{figure*}[t!]
    \centering
    \begin{minipage}{0.44\linewidth}
      \begin{lstlisting}[language=proto,style=mycodestyle]
message LineEntity {
  bool is_construction = 1;
  message Vector {
    double x = 1;
    double y = 2;
  }
  Vector start = 2;
  Vector end = 3;
}
      \end{lstlisting}
    \end{minipage}
    \begin{minipage}{0.44\linewidth}
      \begin{lstlisting}[language=proto,style=mycodestyle]
message MirrorConstraint {
  Pointer mirror = 1;
  message Pair {
    Pointer first = 1;
    Pointer second = 2;
  }
  repeated Pair mirrored_pairs = 2;
}
      \end{lstlisting}
    \end{minipage}%
    \vspace{-6mm}
    \caption{\textbf{Examples of object specifications.} We represent sketch entities and constraints using \emph{Protocol Buffers}. Protocol Buffers allow us to easily write specifications for structured objects of varying complexity. For instance, a line entity is defined by the coordinates of its end points and thus our message has two corresponding fields each of which is a 2D vector. Constraints are specified in a similar fashion except they additionally make use of the \texttt{Pointer} data type. Pointers refer to entities that constraints are applied to.}
    \label{fig:proto_examples}
\end{figure*}

Figure~\ref{fig:proto_examples} shows how we represent the \emph{line entity} and the \emph{mirror constraint} (see Appendix~\ref{sec:appendix_obj_specs} for an extensive list of supported objects). The line specification is straightforward: we first need to decide whether our entity should be treated as a construction geometry (construction geometries will not be displayed in the final drawing) and then provide pairs of coordinates for the beginning and end of the segment. Constraints, such as the mirror constraint, are generally defined in a similar fashion to entities. The \texttt{MirrorConstraint} is used to force an arbitrary number of pairs of geometries (\ie, \texttt{mirrored\_pairs}) to be symmetrical with respect to some axis (\ie, \texttt{mirror}). One may notice that constraints make extensive use of the \texttt{Pointer} data type. Pointer fields are meant to refer to the entities present in the sketch. In practice, we record all eligible pointees (\ie, entities and their parts) in a special table, and a pointer is simply an index of an entry in that table.

\begin{figure}[t]
    \centering
    \begin{lstlisting}[language=proto,style=mycodestyle,escapechar=!]
message Sketch {
  message Entity {
    oneof kind {
      LineEntity line = 1;
      // And other entity types.
    }
  }
  message Constraint {
    oneof kind {
      MirrorConstraint mirror = 1;
      // And other constraint types.
    }
  }
  message Object {
    oneof kind {
      Entity entity = 1;
      Constraint constraint = 2;
    }
  }
  repeated Object objects = 1;
}\end{lstlisting}
    \caption{\textbf{The specification of a full sketch.} A sketch is defined as a sequence of objects. Each object can be either an entity or a constraint. Each entity can be one of the supported types (\eg, a \texttt{LineEntity}). Similarly, a constraint object is resolved to a particular kind of constraint (\eg, a \texttt{MirrorConstraint}).}
    \label{fig:sketch_proto}
\end{figure}

Once all the necessary object types are fully specified we need to convert the data into a form that can be processed by a machine learning model. We choose to represent sketches as sequences of \emph{tokens}. This allows us to pose sketch generation as \emph{language modeling} (LM) and take advantage of the recent progress in this area \citep{radford19,brown20}. To achieve this, we first pack collections of entities and constraints into one Protocol Buffer message in the sketch format defined in Figure~\ref{fig:sketch_proto}. We assume some ordering of objects. For example, we can adopt the ordering provided by the Onshape API, which returns a list for the entities followed by another list for the constraints. The order within each list tends\footnote{Strictly speaking, it is not guaranteed} to reflect the order in which the objects were created by the user. Since constraints refer to entities it makes sense to form the final list of objects in such a way that the beginning of the list is occupied by the entities and the remaining part is allocated for the constraints. In our experiments (see Section~\ref{sec:experiments}), we also explore an alternative ordering in which we insert a constraint right after all the entities it refers to.

There are a few ways to go about obtaining a sequence of tokens from a sketch message. Arguably the most intuitive one is to format messages as text. For a line entity connecting $ (0.0, 0.1) $ and $ (-0.5, 0.2) $ this will result in something similar to:
\begin{lstlisting}[style=mycodestyle]
{
  is_construction: true
  start {
    x: 0.0, y: 0.1
  }
  end {
    x: -0.5, y: 0.2
  }
}
\end{lstlisting}
The text format contains both the structure and the content of the data.  The advantage of using this representation is that we can employ any off-the-shelf approach for modeling textual data. Unfortunately, this comes at a cost: the resulting sequences are prohibitively long even for the modern LM techniques. Additionally, the model would have to generate valid syntax, which would take up some portion of the model's capacity.

We can sidestep some of those limitations by working with \emph{sequences of bytes} forming the \emph{serialized} versions of sketch messages. Serialized PB messages are much shorter since they do not contain any information about the object structure -- this burden is offloaded on to an external \emph{parser} automatically generated from the data specification. The parser's task is to \emph{interpret} the incoming stream of unstructured bytes and populate the fields of PB messages.  However, like the text format, not every sequence of bytes results in a valid PB message.

Going one step further, we can utilize the structure of the sketch format, and build a custom designed interpreter (rather than the generic parser automatically generated from the PB definitions used in the bytes format above), that takes as input a sequence of tokens each representing a valid choice at various decision steps in the sketch creation process.  We designed this interpreter in such a way that all sequences of tokens in this format lead to valid PB messages.

In this format we represent a message as a sequence of \emph{triplets} (we call this the \emph{triplet format}) $ (d_i, c_i, f_i) $ equipped with additional contextual information where $ i $ is an index of the token. The majority of tokens describe basic fields of the sketch objects with each token representing exactly one field. The first two positions in each triplet are allocated for a \emph{discrete value} and a \emph{continuous value} respectively. Since each field in a message is either discrete or continuous only one of two positions is \emph{active} at a time (the other one is set to a default zero value). Ignoring for now the third position in the triplet we can write down the sequence for the line entity above as follows (highlighted are the active positions):
\begin{equation*}
    (\mathbf{1}, 0.0), \, (0, \mathbf{0.0}), \, (0, \mathbf{0.1}), \, (0, \mathbf{-0.5}), \, (0, \mathbf{0.2}) \, .
\end{equation*}
Note that we use the discrete component of the first token to set a boolean field \tokenbox{line.is\_construction}.\footnote{Here and further in the text, we employ the dotted path notation to identify message fields occasionally dropping prefixes where it leads to an unambiguous result.}

Besides basic fields, our sketch specification makes extensive use of the \texttt{\textbf{\textcolor{dmblue400}{oneof}}} construction. In the PB language, \texttt{\textbf{\textcolor{dmblue400}{oneof}}} allows one to introduce mutually exclusive fields. For example, in Figure~\ref{fig:sketch_proto}, we can see that a generic \texttt{Object} can be either an \texttt{entity} or a \texttt{constraint}. To handle this, we simply inject an additional token with the discrete value set to the index of the active field.

Another feature of the PB language that is essential for defining sketches (and many other structured objects) is \texttt{\textbf{\textcolor{dmblue400}{repeated}}} fields. A repeated field is a list of an arbitrary number of elements of the same type. We first see this used in the only top-level field of the \texttt{Sketch} message: a sketch is a list of \texttt{Object}s. In order to represent such fields, we could just concatenate the tokens of all the elements. The only problem we will face is during generation when the model needs to indicate that it has finished producing the list. Specifically for this situation, we use the third position of the token triplet, $ f_i $. It is a \emph{boolean flag} that signifies the end of repetition. The ``end'' token has the main portion set to $ (0, 0.0, \mbox{\textbf{True}}) $.
Note that since messages may have nested repeated fields the resulting sequence may contain several tokens with $ f_i = \mbox{True} $.

Given a sequence of such triplets, it is possible to infer which exact field each token corresponds to. Indeed, the very first token $ (d_1, c_1, f_1) $ is always associated with \tokenbox{objects.kind} since it is the first choice that needs to be made to create a \texttt{Sketch} message (see Figure~\ref{fig:sketch_proto}). The second field depends on the concrete value of $ d_1 $. If $ d_1 =  0 $ then the first object is an \texttt{entity} which means that the second token corresponds to \tokenbox{entity.kind}. The rest of the sequence is associated in a similar fashion. Field identifiers along with their locations within an object form the \emph{context} of the tokens. We use this contextual information as an additional input for our machine learning models since it makes it easier to interpret the meaning of the triplet values and to be aware of the overall structure of the data.

Summarizing the above, a simple sketch with a line entity and a point entity placed at one of its ends will be described with the sequence shown in Table~\ref{tab:seq_example}.
\begin{table}
    \centering
    \begin{tabular}{>{\scriptsize}ll|>{\small}l@{}l}
        & Triplet & Field \\
        \cline{2-3}
        \rule{0pt}{4ex}1. & $ (\mathbf{0}, 0.0, \mbox{False}) $ & \texttt{objects.kind} & \rdelim\}{7}{*}[\,{\rotatebox[origin=c]{-90}{Line}}] \\
        2. & $ (\mathbf{0}, 0.0, \mbox{False}) $ & \texttt{entity.kind} \\
        3. & $ (\mathbf{1}, 0.0, \mbox{False}) $ & \FadeAfter{30pt}{\texttt{line.is\_constr}} \\
        4. & $ (0, \mathbf{0.0}, \mbox{False}) $ & \texttt{line.start.x} \\
        5. & $ (0, \mathbf{0.1}, \mbox{False}) $ & \texttt{line.start.y} \\
        6. & $ (0, \mathbf{-0.5}, \mbox{False}) $ & \texttt{line.end.x} \\
        7. & $ (0, \mathbf{0.2}, \mbox{False}) $ & \texttt{line.end.y} \\
        8. & $ (\mathbf{0}, 0.0, \mbox{False}) $ & \texttt{objects.kind} & \rdelim\}{5}{*}[{\,\rotatebox[origin=c]{-90}{Point}}] \\
        9. & $ (\mathbf{1}, 0.0, \mbox{False}) $ & \texttt{entity.kind} \\
        10. & $ (\mathbf{0}, 0.0, \mbox{False}) $ & \FadeAfter{30pt}{\texttt{point.is\_const}} \\
        11. & $ (0, \mathbf{0.0}, \mbox{False}) $ & \texttt{point.x} \\
        12. & $ (0, \mathbf{0.1}, \mbox{False}) $ & \texttt{point.y} \\
        13. & $ (0, 0.0, \mbox{\textbf{True}}) $ & \texttt{objects.kind} \\
    \end{tabular}
    \caption{\textbf{A triplet representation of a simple sketch.} The sketch contains and a line and a point. Within each triplet in the \emph{left column}, the \emph{active} component (the value that is actually used) is highlighted in \textbf{bold}. The \emph{right column} shows which field of the object the triplet should be associated with.}
    \label{tab:seq_example}
\end{table}

\section{Model}
\label{sec:model}
Our main goal is to estimate the distribution $ \pdata $ of 2D sketches in a dataset $ \data $.  As described in the previous section, we are going to treat a sketch as a sequence of tokens. In this work, we only consider the byte and the triplet representations due to the sequence length challenges associated with the raw textual format.

In both cases, we decompose the joint distribution over the sequence of tokens $ \mathbf{t} = (t_1, \ldots, t_N) $ as a product of conditional distributions:
\begin{equation}
\label{eq:prod_of_conds}
    p(\vt; \theta) = \prod_{i = 1}^{N} p(t_i \mid t_{<i} \, ; \theta) \, ,
\end{equation}
where $ N $ is the length of the sequence and $ t_{<i} $ denotes all the tokens preceding $ t_i $. Following the standard approach for modeling sequential data \citep{mikolov10}, we employ an autoregressive neural network parameterized by $ \theta $ to obtain $ p(t_i \mid t_{<i} \, ; \theta) \; \forall \, i $. The task of $ \pdata $ estimation becomes maximization of the log-likelihood of $ \data $ under the model, \ie,
\begin{equation}\label{eq:max_likelihood}
    \theta^{*} = \arg\max_{\theta} \sum_{\vt \in \calD} \log p(\vt; \theta) \, .
\end{equation}

More concretely, we employ the \emph{Transformer decoder} architecture \citep{vaswani17} that accepts a vector representation of a token $ \ve_{i - 1} = \mbox{\texttt{embed}}_{\,i}(t_{i - 1}) \in \mathbb{R}^D $ (an \emph{embedding}) and maps it into another vector $ \vh_i $ of the same dimensionality. The latter is then decoded into parameters of $ p(t_i \mid t_{<i}) $ by means of, for example, a linear projection $ \mbox{\texttt{proj}}_{i}(\cdot) $.

\subsection{Byte representation}
\label{sec:byte_model}

When dealing with the bytes of a PB message, each token is simply a discrete value in the range $ \{ 0, \ldots, 255 \} \cup \{ \mathtt{EOS} \} $\footnote{$ \mathtt{EOS} $ denotes the end of the sequence.} and therefore $ p(t_i \mid t_{<i} \, ; \theta) $ can be modeled as a \emph{categorical} distribution similar to how it's done in typical LM approaches \citep{bengio03}. In this setting, for each time step $ i $ of the sequence we have
\begin{equation}
\label{eq:byte_embed}
    \mbox{\texttt{embed}}_{\,i}(t_{i - 1}) = V[t_{i - 1}] + \ve_i^{\mbox{\footnotesize pos}} \, ,
\end{equation}
where $ [\cdot] $ denotes the lookup operation and $ \ve_i^{\mbox{\footnotesize pos}} $ is a position embedding for position $i$. Both $ V $ and $ \ve_i^{\mbox{\footnotesize pos}} $ are learned. Moreover, $ \forall i \; \mbox{\texttt{proj}}_{i} \equiv \mbox{\texttt{proj}} : \mathbb{R}^D \ra \mathbb{R}^{257} $ and the output is treated as \emph{logits} of the distribution.

\subsection{Triplet representation}
\label{sec:triplet_model}

In case of the triplet representation, we follow a slightly more involved procedure. As outlined in Section~\ref{sec:data_layout}, tokens can be either discrete or continuous. Additionally, different discrete tokens may have \emph{different ranges of values}. For example, there are only two possible values for the \tokenbox{object.kind} token -- either to an entity or a constraint. On the other hand, the range of the \tokenbox{entity.kind} token has cardinality of 4 since we support 4 different types of sketch entities. This means that we can't naively describe each conditional in Equation~\ref{eq:prod_of_conds} using the same template distribution like we did for the bytes. We circumvent this by introducing the notion of \emph{token groups}.

A token group $ \calG $ is a collection of related token types that can be handled in a similar fashion. Specifically, we use the same embedding function and the same projection for each $ t \in \{ t \mid \mbox{\texttt{type}}(t) \in \calG \}  $. For instance, we might want to group all the tokens associated with coordinates. In the example from Table~\ref{tab:seq_example}, tokens \tokenbox{line.start.x}, \tokenbox{line.start.y}, \tokenbox{line.end.x}, \tokenbox{line.end.y}, \tokenbox{point.x} and \tokenbox{point.y} will all end up in the same $ \calG $.\footnote{For the full list of token groups used in the model see Appendix~\ref{sec:appendix_token_groups}.} Whenever we need to compute the log-likelihood of those tokens in the sequence we can employ the same functional form of the output distribution, \eg, a mixture of Gaussians \citep{graves13,ha17} with the same number of components. In practice, however, we replace continuous values with their 8-bit uniformly quantized versions and model everything using categorical distributions since we found it to greatly improve the stability of training and the overall performance. This is in line with the observations made in \citep{nash20,oord16}.

We embed $ t_{i - 1} $ that belongs to the group $ \calG $ as (note the difference with Equation~\ref{eq:byte_embed}):
\begin{equation}
\label{eq:triplet_embed}
    K[\mathtt{field}(t_{i - 1})] + V^{\calG}[t_{i - 1}] + \ve_n^{\mbox{\footnotesize obj}} + \ve_m^{\mbox{\footnotesize rel}} \, ,
\end{equation}
where $ \mathtt{field}(t) $ returns the categorical field identifier associated with $ t $ and $ K $ is a collection of learnable embeddings for every possible field type. Unlike in Equation~\ref{eq:byte_embed}, instead of using global position embedding $ \ve_i^{\mbox{\footnotesize pos}} $ we describe the location with the index $ n $ of the current object as well as the relative position $ m $ of $ t_{i - 1} $ within an object.

We handle the ``end'' tokens (\ie, $ f_i = \mbox{True} $) similarly to $ \mathtt{EOS} $ in Section~\ref{sec:byte_model} -- the output projection produces an additional logit used to compute the probability of ending the repetition. Since $ f_i $ is only expected to be $ \mbox{True} $ at particular points in the sequence (\ie, right after tokens forming a \emph{whole} element of the list) we \emph{mask out} the extra logit everywhere else. This ensures that the ``end'' token can't be predicted prematurely and also eliminates its unnecessary contribution to the optimized objective.

One significant difference between the byte setting and the triplet setting is how we process pointer fields. In the former, pointers do not get any special treatment and are generated just like any other integer field. We rely on the model's capability to make sense of the entity part index and relate it to the corresponding locations in the sequence via attention weights. This seems to be a viable strategy since Transformers have demonstrated an impressive referencing capacity in recent works \citep{brown20}.

Since the triplet representation provides us with direct access to the semantics of tokens it's possible to relate pointers to their pointees more explicitly by using Pointer Networks \citep{vinyals15}. The approach we are taking here is similar to \citep{nash20}. In order to compute $ p(t_i \mid t_{<i} \, ; \theta) $, we first project the output of the Transformer $ \vh_i $ into the final pointer vector $ \vp_i = W_{\mbox{\footnotesize ptr}} \, \vh_i $. The conditional is then obtained as follows:
\begin{gather}
    p(t_i \; \mbox{\small points to} \; t_{ik} \mid t_{<i} \, ; \theta) = \softmax_k (\vp_i^T H_i) \, , \\
    H_i = \left[ \vh_{i1}, \ldots, \vh_{ik}, \ldots, \vh_{iM} \right] \, ,
    \label{eq:transformer_output}
\end{gather}
where $ \softmax_k $ is the $ k $-th element of the softmax vector and $ \vr_i = \{ t_{i1}, \ldots, t_{iM} \} $ is the set of tokens that $ t_i $ can point to. Naturally, $ \vr_i $ only contains tokens from $ t_{<i} $. This is different from \citep{vinyals15,nash20} where $ \vr_i \equiv \vr $ is external to the predicted sequence and remains immutable throughout the generation process.

The pointer mechanism takes into account that there is no clear correspondence between the tokens that describe an entity and the entity's parts that pointers can refer to in the data. For example, one may want to create a constraint that applies to the midpoint of a line segment. Let's imagine\footnote{Hypothetically, since Onshape does not operate with midpoints as object parts. A real example would involve splines and their endpoints which are harder to use as an illustration.} that the Onshape API exposed the midpoint as an object part. Since the line is only defined by its endpoints (see Figure~\ref{fig:sketch_proto}, lines 1 - 7) there is no obvious candidate token that we could refer to in the constraint definition. For that reason, we introduce special \emph{referrable tokens} decoupling pointers from the object definition tokens. Referrables are injected after each entity and have the same identifier within each entity type. In order to let the model distinguish between different entity parts, we add the index of the referrable to the embedding of the token. Continuing with the line segment example and assuming that the entity has 4 parts, \ie, the line as a whole (\textbf{part~1}), 2 endpoints (\textbf{parts~2} and \textbf{3}), and the midpoint (\textbf{part~4}), the token $ t_i $ for the midpoint will have the following embedding:
\begin{equation}
    K[\mbox{\tokenbox{line.ref}}] + V^{\mbox{\footnotesize ref}}[4] + \ve_i^{\mbox{\footnotesize pos}} ,
\end{equation}
where $ V^{\mbox{\footnotesize ref}} $ is a learnable vocabulary. As referrable tokens do not need to be predicted the respective terms are removed from Equation~\ref{eq:prod_of_conds}. For the same reason, in Equation~\ref{eq:transformer_output}, $ \vh_{ik} $ corresponds to the time step where $ t_{ik} $ is the \emph{output} and not the input. This allows us to avoid having unused Transformer outputs.

Finally, we need to specify how we embed pointers as inputs to the Transformer network. Following \citep{vinyals15,nash20} we could reuse $ \vh_j $ for tokens that point to $ t_j $. Unfortunately, this creates output-to-input connections which are extremely detrimental to the efficiency of the Transformer architecture --  different time steps can no longer be processed in parallel during training. Instead, we opt for a simpler solution and employ the standard embedding scheme for discrete tokens (Equation~\ref{eq:triplet_embed}).

\begin{figure*}[p]
    \centering
    \begin{subfigure}[b]{\textwidth}
        \centering
        \includegraphics[width=0.95\linewidth]{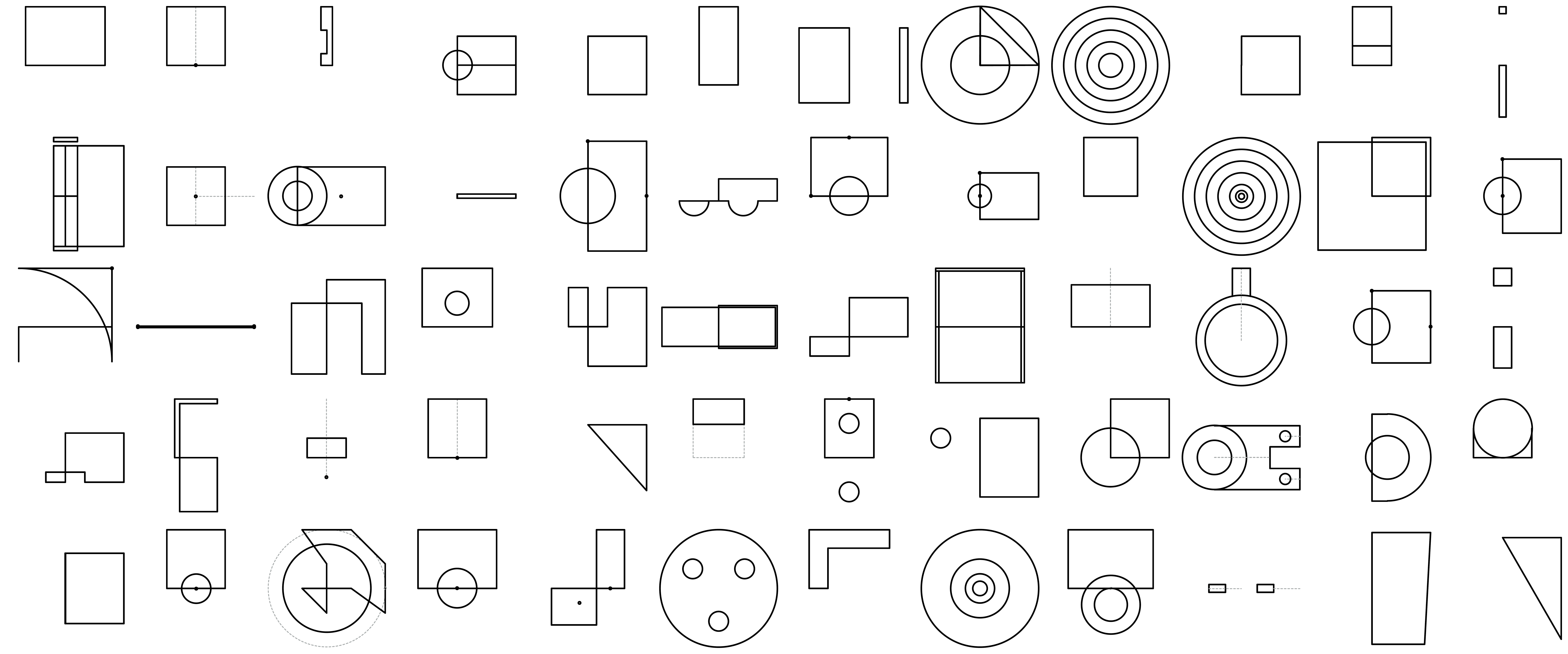}
        \caption{\textbf{Unconditional byte} model. }
        \label{fig:byte_uncond}
    \end{subfigure}
    \\[4mm]
    \begin{subfigure}[b]{\textwidth}
        \centering
        \includegraphics[width=0.95\linewidth]{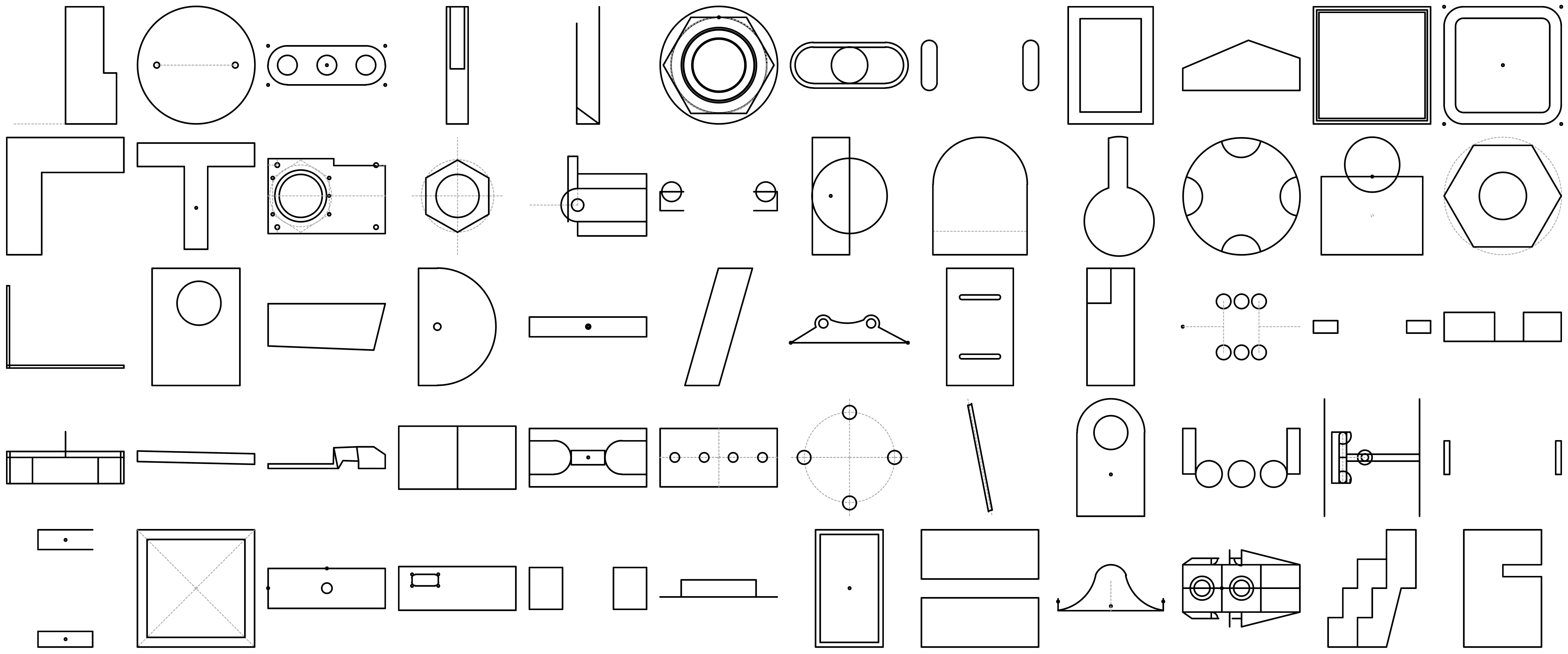}
        \caption{\textbf{Unconditional triplet} model.}
        \label{fig:tri_uncond}
    \end{subfigure}
    \\[4mm]
    \begin{subfigure}[b]{\textwidth}
        \centering
        \input{figures/samples/tri_cond}%
        \caption{\textbf{Conditional triplet} model. Model predictions are rendered using a slightly thicker style.}
        \label{fig:tri_cond}
    \end{subfigure}
    \caption{\textbf{Random samples} from various models. We use Nucleus Sampling with \textit{top-p} $ = 0.9 $.}%
\end{figure*}

\subsection{Sampling from the model}

Sampling from the byte model is straightforward. The process is identical to any typical Transformer-based LM. The triplet model, on the other hand, requires slightly more bespoke handling. Figure~\ref{fig:main} illustrates the procedure. We start by embedding and feeding a special $ \mathtt{BOS} $ token into the Transformer. The Transformer then outputs a collection of triplets, one for each possible token group. In order to determine which concrete token needs to be emitted, we employ an interpreter (a state machine) automatically generated from the data specification. Knowing the current state allows us to choose the right token group and associate the active component of the triplet with a field in the synthesized object. Once the appropriate field is populated the interpreter transitions to the next state and produces an output token which is then fed back into the model. The process stops when the state machine receives the ``end'' triplet for the outermost repeated field (\ie, \tokenbox{object.kind}). 

\subsection{Conditional generation}
\label{sec:conditional_model}

In addition to the model described above that generates sketches from scratch, we explore a modification that allows generation of a sketch based on its visual render or drawing.
The conditional model relies on the same transformer-based architecture with an additional input sequence $H^{\text{img}} = [ \mathbf{h}_{1}^{\text{img}}, \ldots, \mathbf{h}_{P}^{\text{img}} ]$ to which the transformer can attend at every layer. 
We obtain $H^{\text{img}}$ by embedding the conditioning image using visual transformer~\citep{dosovitskiy2020} which extracts patches of a certain size from the image and processes them by a number of self-attention layers.

This scheme not only allowed us to conveniently contain the conditional model within the same transformer framework, but also allowed the generator to attend only to relevant parts of the image when processing each of the objects.
This proved to be crucial for reconstructing finer details of the sketch and could not be achieved with a more traditional single-vector representation of the image. 

We train the conditional model in a manner similar to~\eqref{eq:max_likelihood}:
\begin{equation}
    \theta^* = \arg \max_{\theta} \sum_{t \in \mathcal{D}} \log p(\mathbf{t} | \text{image}(\mathbf{t}); \theta),
\end{equation}
where $\text{image}(\mathbf{t})$ is a computer render for the token sequence $\mathbf{t}$ and $\theta$ now includes parameters of the visual transformer too.
Whenever a parallel dataset is available consisting of sketches and their drawings (e.g. human-drawn), the model can be trained accordingly which we leave for future work. 

\section{Experiments}
\label{sec:experiments}
\begin{figure*}[p]
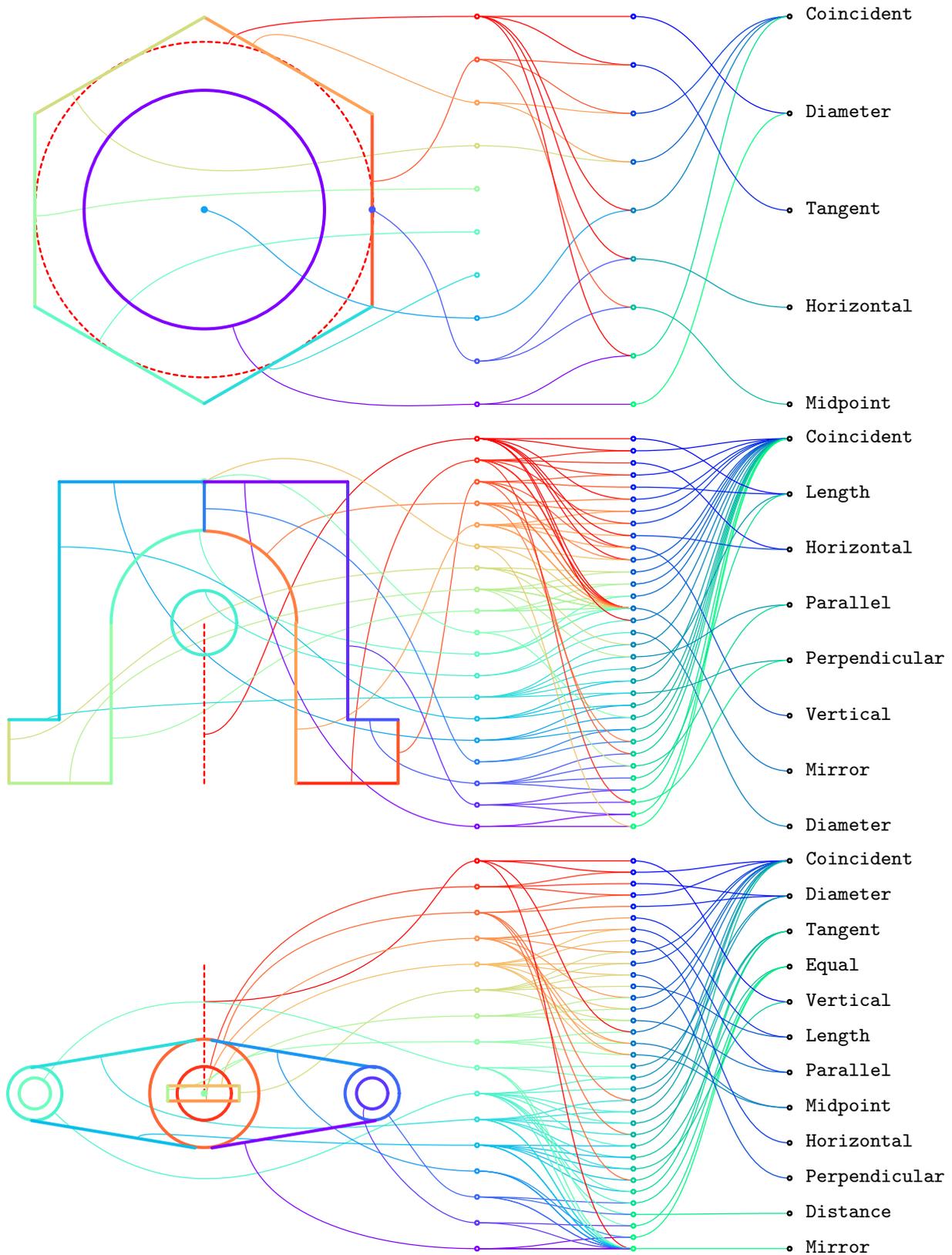

    \centering
    \input{figures/samples/uber_2}\\
    \input{figures/samples/uber_3}\\
    \input{figures/samples/uber_1}\\
    \caption{\textbf{Entities and constraints} sampled from the triplet model. The \emph{first column} of nodes represents different entities. The order of nodes (top to bottom) follows the generation order. The \emph{second column} represents different constraints also ordered by their index in the sequence. Finally, the \emph{third column} is reserved for constraint types, from the most to the least frequent.}%
    \label{fig:uber}%
\vspace{-9pt}
\end{figure*}

\begin{figure*}[t]
    \centering
    \input{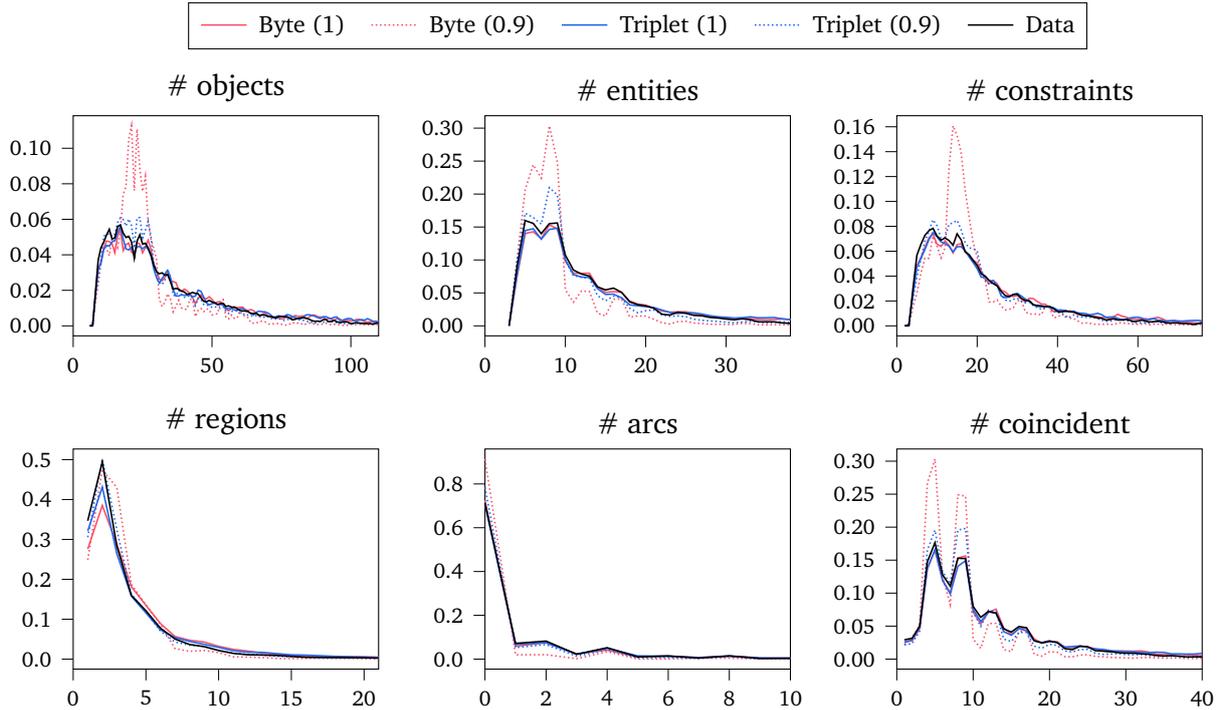}%
    \caption{\textbf{Distribution of various sketch statistics} for samples drawn from our unconditional models. The \textit{top-p} parameter for Nucleus Sampling is shown in parentheses.}
    \label{fig:stats_main}
\end{figure*}

We validate our proposed approaches on the data that we obtained from the repository of documents publicly available on the Onshape platform \citep{onshape}. Following the standard evaluation methodology for autoregressive generative models \citep{oord16,nash20} we use log-likelihood as our primary quantitative metric. Additionally, we provide a variety of random and selected model samples for qualitative assessment.

\subsection{Dataset}

The data we rely on in the work originates from the Onshape website. The website contains a big collection of CAD models which can be used freely for non-commercial purposes. Because of this there have been several attempts to employ Onshape as a source of research data \citep{koch19,seff20}. Unfortunately, in the context of CAD sketches, all the available datasets derived from the Onshape public repository seems to have limitations. In particular, the Onshape API makes it difficult to obtain raw parameters of sketch entities and constraints since they are often defined as symbolic expressions with variables. These variables may in turn correspond to expressions that are specified elsewhere in the document and so on. The simplest way to deal with this is to just ignore such sketches and only consider instances relying on plain values. Prior efforts take this route and therefore discard a significant portion of the available data.

\begin{table}[t!]
    \renewcommand{\arraystretch}{1.5}
    \centering
    \begin{tabular}{llcc}
        \hline
        \multirow{2}{*}{\textbf{Model}} & \multirow{2}{*}{\textbf{Sequence}} & \multicolumn{2}{c}{\textbf{Average bits per}} \\\cline{3-4}
        & & \textbf{object} & \textbf{sketch} \\
        \hline
        Uniform & \small{unord. bytes} & $ 112.23 $ & $ 3683.56 $ \\
        & \small{unord. triplets} & $ 25.34 $ & $ 847.52 $ \\
        Byte & \small{concatenated} & $ 4.394 $ & $ 133.281 $ \\
        & \small{interleaved} & $ 4.265 $ & $ 128.053 $ \\
        Triplet & \small{concatenated} & $ 4.214 $ & $ 127.607 $ \\
        & \small{interleaved} & $ 4.092 $ & $ 122.818 $ \\
        Cond. & \small{interleaved} & $ 2.620 $ & $ 79.979 $ \\
    \end{tabular}
    \caption{\textbf{Test likelihoods of various models.} The \emph{third column} is computed as a mean number of bits per object in a sketch averaged across test examples. The \emph{fourth column} is similar except we do not divide by the number of objects.}
    \label{tab:ll}
\end{table}

In order to collect a more complete dataset of sketches, we performed a large-scale scrape downloading around 6TB of public documents in October, 2020. We only consider sketches containing the most common entity and constraint types (4 and 16 different types respectively). Additionally, we remove constraints that refer to so-called ``external'' entities, \ie, elements of the CAD model that exist outside the sketch plane.

We then preprocess the remaining data to replace symbolic expressions with plain values. To that end, we first employ the \emph{default configuration} file to obtain the initial assignments of variables and then scan through the document locating further assignments and updating the variable mapping. If an assignment contains a symbolic expression we evaluate it using SymPy\footnote{This works most of the time but there are rare cases when SymPy fails to parse Onshape's syntax.} \citep{meurer17}. Once the scan is finished we can use plain values of the variables to resolve any expressions in the sketch parameters.

CAD models in the public repository cover a wide range of objects in the real world, anything from coffee mugs to sports cars. As a result, sketches have significant variability in size. To make training of our models easier, we rescale examples in the dataset to lie in the $ [-1, 1] \times [-1, 1] $ bounding box.

Another topic that is somewhat overlooked in \citep{seff20,koch19} is duplicate data. As is the dataset ends up having many sketches that are either single built-in primitives or copied (sometimes with minor modifications) from popular documents or tutorials. Not only this restricts the diversity of samples generated by machine learning models but also makes assessment of such models difficult -- a random subset chosen for testing is likely to significantly overlap with the training set.

We make a best-effort attempt to address duplication by using the following filtering procedure. We first remove simple axis-aligned rectangles which constitute around $ \approx 15\% $ of the data. Next, we divide the remaining examples into bins of \emph{semantically equivalent} sketches. Two sketches are considered to be equivalent if they share the same sequence of \emph{object types} (\eg, a line followed by a point followed by a coincidence constraint). For more complex examples that have matching sequences of types there is a good chance that the examples themselves look very similar.

The resulting bins contain manageable numbers of sketches (up to several tens of thousands) and therefore can undergo more computationally expensive filtering. Within each bin, we obtain $ 128 \times 128 $ binary renders of each example and apply hierarchical clustering\footnote{We use \texttt{cluster.hierarchy.fclusterdata} from SciPy.} with $ 0.1 $ threshold and Jaccard distance as metric. The final dataset is formed by taking a single sketch per cluster (unsurprisingly, sketches from the same cluster end up looking almost identical).

For our experiments, we exclude the sketches that have fewer than 4 or more than 100 entities. The dataset is split randomly into 3 parts: \num{4671037} examples for the training set and \num{50000} sketches for each the validation and the test set.

\subsection{Training details}

We train our models for \num[retain-unity-mantissa=false]{1e6} weight updates using batches with 128 lanes. Each lane can fit sequences up to 1024 tokens long in the triplet setting and 1990 tokens long in the byte setting. In order to improve occupancy and reduce wasted computation we fill up the lanes dynamically packing as many examples as possible into a lane before moving on to the next one. Each batch is processed in parallel by 32 TPU cores.

We use the Adam optimizer with the learning rate of \num[retain-unity-mantissa=false]{1e-4} and clip the gradient norm to $ 1.0 $. Additionally, we employ a dropout rate of $ 0.1 $ in all the experiments.

\subsection{Unconditional generation}
\label{sec:uncond_generation}

In this series of experiments, the goal is to determine how well our models capture the distribution of sketches in the dataset. We use the same network architecture for both the byte and the triplet settings: embeddings and fully-connected layers of size $ 384 $ and $ 24 $ Transformer blocks. For the baseline, we employ a model that outputs uniform distribution at every time step.

As we mentioned in Section~\ref{sec:data_layout}, we have to choose the ordering of objects in training sequences since both entities and constraints are given to us as sets. We compare two regimes: in the first one (\emph{concatenated}), constraint objects go after the last entity while in the second (\emph{interleaved}), a constraint object is injected immediately after the entities it operates on.

Table~\ref{tab:ll} shows test log-likelihoods obtained by different models. Unsurprisingly, our proposed methods (the last 4 rows) significantly outperform the weak baselines. The difference between the two uniform settings is due to the fact that the byte description of a PB message is usually longer than the triplet one: 239 \versus 456 tokens on average with the maximum length of 959 \versus 1987. Additionally, the tokens in the triplet representation tend to have smaller cardinality (\ie, $ < 257 $).

These differences between representations may partially explain why triplet models demonstrate better performance on the hold-out test set. It's also worth emphasizing that the byte model does not receive any explicit information about the parsing state. This seem to make learning more challenging and as a result compared to the triplet model it takes roughly 3 times more network updates to reach the highest data likelihood on the validation.

We tried making byte sequences shorter by feeding PB messages into a general purpose compression algorithm (Brotli \citep{alakuijala18} with the quality setting of 7). This allowed us to reduce the maximum length to 930 tokens, which is very similar to that of the triplet setting. Unfortunately, this change renders learning impossible -- a model of the same capacity as before fails to improve beyond the initial reduction of the training objective. We suspect that this happens because the compression algorithm aggressively mangles the contents of a PB message and thus it is too difficult for the model to make sense of the data.

Another important factor affecting the performance of the models is the choice of the object ordering. As it is evident from Table~\ref{tab:ll}, the interleaved ordering consistently leads to better results. One explanation for this is that at any point in time, the model has more explicit information about the relations between the sketch entities produced so far. This early injection of the design intent seems to make it easier to decide what to generate next.

In addition to measuring likelihoods, we sampled \num{10000} sketches from the best performing byte and triplet models and computed distributions of various high-level statistics (\eg, number of objects, number of closed regions and number of coincident constraints). We repeated this procedure both with and without using Nucleus Sampling \citep{holtzman19}. Figure~\ref{fig:stats_main} (as well as Figure~\ref{fig:stats_appendix} in the appendix) reveals that both models follow the data distribution closely when we use samples from the unmodified model output. In this setting, however, a significant fraction of sketches is either malformed (\eg, the generated PB message cannot be parsed) or unsolvable: $ 36 \% $ for the byte model and $ 14 \% $ for the triplet model. Nucleus Sampling with \textit{top-p} $ = 0.9 $ skews the sample distribution and seems to have a more pronounced negative effect on the byte model. The upside is that the resulting sketches become ``cleaner'': the percentage of invalid samples goes down to $ 25 \% $ and $ 6 \% $ for the byte and the triplet settings respectively.

Figures~\ref{fig:byte_uncond} and \ref{fig:tri_uncond} show renders of random samples from several proposed models. Overall, generated sketches look plausible and exhibit a lot of desired properties: closedness of regions, regularity, symmetry, a non-trivial amount of fine detail. We observe that the byte model tends to produce slightly less complex samples with fewer open arcs but this could be a side effect of applying  Nucleus Sampling with the chosen parameter. We also note that the model does not always synthesize sensible sketches -- just like any other typical autoregressive model trained with \emph{teacher forcing} it suffers from not being able to recover from mistakes made early on in the sequence. This can potentially be addressed by fine-tuning using, for example, reinforcement learning.

\begin{figure}[t]
    \centering
    \includegraphics[width=0.95\linewidth]{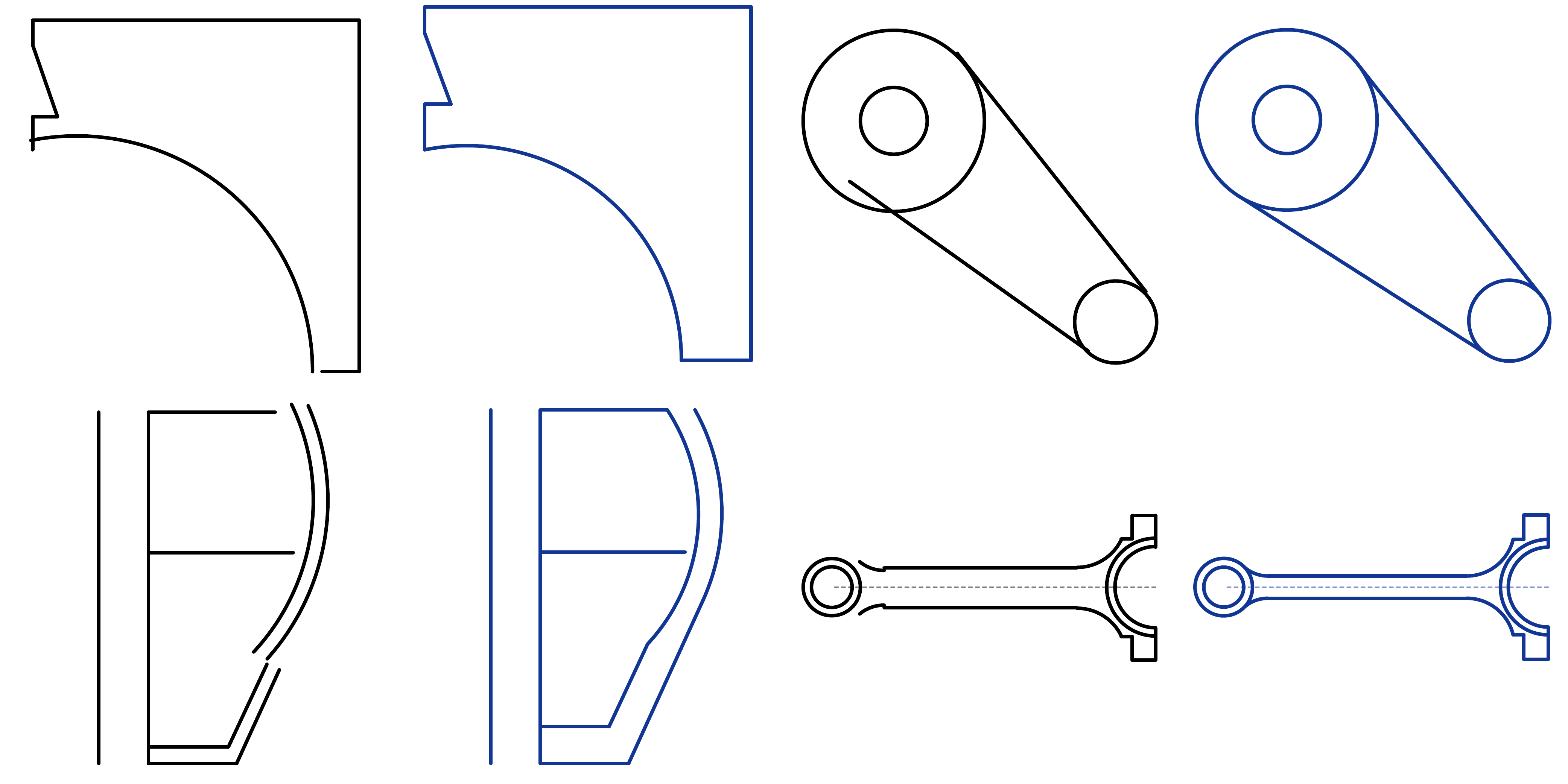}
    \caption{\textbf{Examples of solved sketches.} Our models predict both entities and constraints and thus we can run samples through a dedicated solver that adjusts the parameters of the entities to respect the constraints. This process can fix imperfections in the initial configuration (rendered in \emph{black}).}
    \label{fig:solved}
\end{figure}

\begin{figure*}[t!]
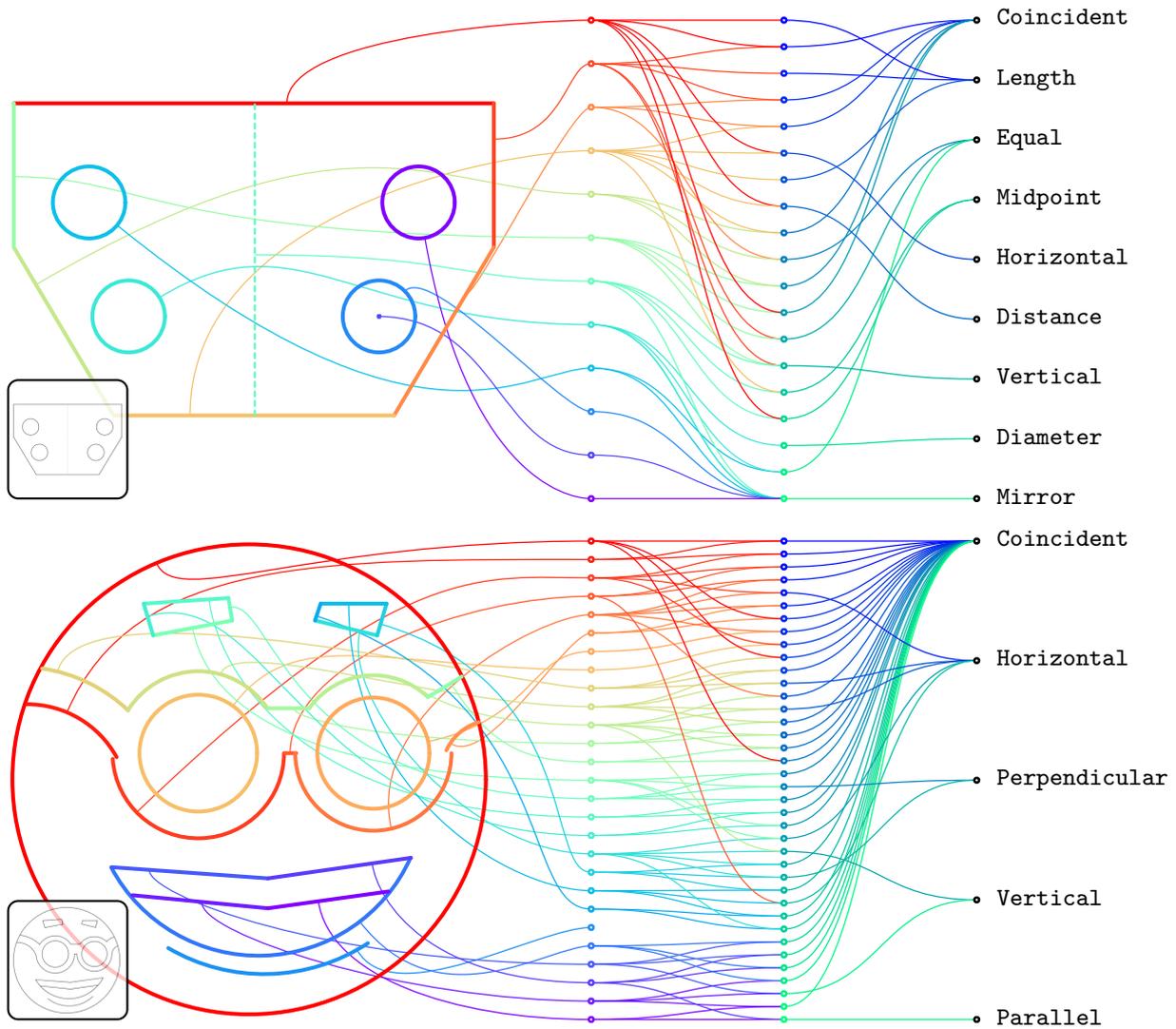

    \centering
    \input{figures/samples/cond_uber}\\
    \input{figures/samples/cond_uber_cool}\\
    \caption{\textbf{Entities and constraints inferred by the conditional model.} Input bitmaps are shown in the lower left corner. The bottom example demonstrates the model's performance on an out-of-distribution input. See Figure~\ref{fig:uber} for further explanation.}%
    \label{fig:cond_uber}%
\end{figure*}

In order to make visualization of sketch constraints more manageable, we show several additional scaled up samples from the triplet model in Figure~\ref{fig:uber}.\footnote{Here, we do not distinguish between different parts of an entity and fold all of them into a single node in the first column.} While not perfect, the inferred constraints are reasonable most of the time. The model does a good job at connecting entities using \texttt{Coincident} constraints (perhaps unsurprisingly since it's the most common type in the data) but also successfully detects more complex relations spanning more than two primitives (\eg, \texttt{Mirror}). Having access to constraints gives us opportunity to correct mistakes in entity prediction by applying an external sketch solver (see Figure~\ref{fig:solved}). Although this aspect of sketching was not the main focus of this work, we believe that a tighter integration between the model and the CAD software will lead to a significant boost in generation quality.

\subsection{Conditional generation}

As discussed in section~\ref{sec:conditional_model}, we also trained an image-conditional model using the same regime as for the main model.
We used 16 and 22 layers for the visual and triplet transformers correspondingly, both operating with 384-dimensional embeddings.
We employed $256 \times 256$ sized renders and $16 \times 16$ image patches resulting in total $P=256$ conditioning vectors $\mathbf{h}_p^{\text{img}}$.

As can be seen in table~\ref{tab:ll}, the conditional model (\ie, \emph{Cond.}) achieves a significantly better fit than the unconditional variant. One can notice, that the model has still retained a non-trivial amount of uncertainty arising from the fact there can be many different token sequences resulting in the same image (e.g. due to different permutations), so the reverse transformation is non-deterministic. 

Image-conditional samples can be found on Figure~\ref{fig:tri_cond}. The model was able to nearly perfectly reconstruct simpler sketches and mostly made mistakes in the presence of a large number of fine details. 

We were also interested if the image-conditional model has learned a meaningful algorithm for CAD conversion that would generalize to images that are implementable within our sketch vocabulary but come from a different distribution of objects. Thus, we drew a smiley face in Onshape and processed the image using our model.
We discovered that the model is able to very sensibly convert that image, but unfortunately decides to stop the generation process too early due to the fact that the majority of training sequences contained much smaller number of examples than used in the smiley face.

To overcome this arguably insignificant limitation we enforced the interpreter to continue generating tokens until the limit of 1024 tokens is reached. 
We then selected the longest sub-sequence from 64 sampled sequences that minimized the $L_2$ difference between the conditioning image and rendered result (after Gaussian smoothing with $\sigma=4$), which can be seen as a form of model-guided search procedure.
The resulting nearly perfect sketch can be seen on Figure~\ref{fig:cond_uber}. 

One can note than even though the model did not place the arcs from which the eye-glasses are built very precisely, it still supplemented them with constraints that make sure their endpoints coincide. Besides this, mouth-forming lines drawn as almost parallel in the original sketch simply by the chance, were recognized as parallel in the provided sample with the corresponding constraints which is an example of good understanding of relationships between objects by the model.

\section{Discussion}
In this work, we have demonstrated how a combination of a general-purpose language modeling technique alongside an off-the-shelf data serialization protocol can be used to effectively solve generation of complex structured objects. We showcased the proposed system on the domain of 2D CAD drawings and developed models that can synthesize geometric primitives and relations between them both unconditionally as well as using a bitmap as a reference. These are only initial proof-of-concept experiments and we are hoping to see more applications taking advantage of the flexibility of the developed interface: conditioning on various sketch properties, inferring constraints given entities and automatically completing drawings, to name a few.

Although we focused our attention on a particular dataset we argue that the approach described in this paper is largely domain-agnostic. In order to adapt the system to a new kind of data, the algorithm designer only needs to provide an appropriate Protocol Buffer specification and if the PB language is too restrictive one can always replace it with a more powerful interpreter. As a straightforward direction for future work, we can consider extending the method to handle 3D. In Onshape, 3D operations bear a lot of similarities with sketch constraints -- just like constraints, they can be represented as nested messages containing references to the geometries existing in the scene (\eg, an extrusion operation refers to a 2D profile to be extruded). Thus, most of the ideas from the present work can be taken verbatim to this new setting.

When it comes to model architecture, there are a few ways in which we can potentially improve it. For example, we can enhance the input fed into the Transformer at every time step by incorporating a visual representation of a partially constructed sketch. There is evidence that this may lead to better performance \citep{ganin18} as the model gains more direct access to the holistic state of generation. Another improvement we briefly mentioned in Section~\ref{sec:uncond_generation} is tighter interaction between the model and the CAD software which makes ultimate use of the synthesized data. For instance, currently, our training procedure does not take into account how the sketch solver would react to adding more entities and constraints. The model is simply trying to imitate valid examples from the dataset. As a result, the model's capacity is not always utilized efficiently: a lot of it may be allocated on capturing the minutiae of sketch parameters with more important properties like solvability and stability being overlooked. Therefore it makes sense to introduce training objectives that are directly correlated with the desired behaviour of the system. This brings us to the territory of reinforcement learning with rich feedback from the software environment, a setting which is much closer to how sketching is done by human designers.

We hope that this work will serve as a stepping stone for further advances in the field of automated CAD but also inspire new ideas and approaches to generative modeling of arbitrary structured data.

\bibliographystyle{abbrvnat}
\nobibliography*
\bibliography{report}

\section*{Acknowledgements}
The authors would like to thank Charlie Nash and Georg Ostrovski for helping with the manuscript preparation as well as Igor Babuschkin, David Choi, Nate Kushman, Andrew Kimpton, Jake Rosenfeld, Lana Saksonov, John Rousseau, Greg Guarriello, Andy Brock, Francesco Nori, A\"{a}ron van den Oord and Oriol Vinyals for insightful discussions and support. 

\section*{Author Contributions}
YG led the project, designed the data layout and the triplet model, collected the dataset, implemented and conducted the bulk of the experiments, wrote the core of the manuscript. SB implemented and evaluated the conditional model. YL contributed the byte model. EK provided technical support. YG, SB, YL, EK and SS worked on editing the paper. 

\clearpage
\appendix
\section{Object specifications}
\label{sec:appendix_obj_specs}

This section contains additional details on the object specifications. As mentioned in Section~\ref{sec:data_layout}, we rely on the PB language to define the structure for each object type that we would like to handle with our model. Our framework supports all basic constructions on the language including nested messages, \texttt{\textbf{\textcolor{dmblue400}{repeated}}} fields and \texttt{\textbf{\textcolor{dmblue400}{oneof}}} clauses. The latter is especially useful when a particular object may appear in several mutually exclusive configurations: for example, \texttt{CircleArcEntity} represents both arcs and closed circles (for which it does not make sense to specify end points). Sometimes the configuration is determined by a group of preceding field values. In that case, the \texttt{\textbf{\textcolor{dmblue400}{oneof}}} branch is chosen by a special \emph{handler} that observes the fields populated so far. This handler is passed as an \emph{option} to the construction (see \texttt{DistanceConstraint} definition below). Options provide a convenient way to extend the language and make it better suited for a domain of interest. For example, we use the \texttt{at\_least} field option to indicate the minimum number of items in a \texttt{\textbf{\textcolor{dmblue400}{repeated}}} field.

Below we are listing all the supported object specifications. Although most of them are direct adaptations of the JSON structures returned by the Onshape API we made an effort to remove redundant and irrelevant sections. We also attempted to bring our constraint specifications closer to how they behave in the software UI. For instance, the \emph{coincident constraint} can usually be applied to several entities at once. In the API, however, this is translated into a collection of pairwise constraints. In our dataset, we compress this collection back into a single object. 

We start with the \emph{sketch entities}:
\begin{lstlisting}[language=proto,style=mycodestyle,escapeinside=``]
message Vector {
  double x = 1;
  double y = 2;
}

message PointEntity {
  bool is_construction = 1;
  Vector point = 2;
}

`\pagebreak`
`\\\vspace{-3\baselineskip}`

message LineEntity {
  bool is_construction = 1;
  Vector start = 2;
  Vector end = 3;
}

message CircleArcEntity {
  bool is_construction = 1;
  Vector center = 2;
  message CircleParams {
    double radius = 1;
  }
  message ArcParams {
    Vector start = 1;
    Vector end = 2;
    bool is_clockwise = 3;
  }
  oneof additional_params {
    CircleParams circle_params = 3;
    ArcParams arc_params = 4;
  }
}

message InterpolatedSplineEntity {
  bool is_construction = 1;
  bool is_periodic = 2;
  repeated Vector interp_points = 3
      [(field_options).at_least = 2];
  Vector start_derivative = 4;
  Vector end_derivative = 5;
  message TrimmedParams {
    double start_phi = 1;
    double end_phi = 2;
  }
  oneof additional_params {
    Empty untrimmed_params = 6;
    TrimmedParams trimmed_params = 7;
  }
}
\end{lstlisting}

The \emph{constraints} are specified are as follows:
\begin{lstlisting}[language=proto,style=mycodestyle]
message FixConstraint {
  repeated Pointer entities = 1
      [(field_options).at_least = 1];
}

message CoincidentConstraint {
  repeated Pointer entities = 1
      [(field_options).at_least = 2];
}

message ConcentricConstraint {
  repeated Pointer entities = 1
      [(field_options).at_least = 2];
}

message EqualConstraint {
  repeated Pointer entities = 1
      [(field_options).at_least = 2];
}

message ParallelConstraint {
  repeated Pointer entities = 1
      [(field_options).at_least = 2];
}

message TangentConstraint {
  Pointer first = 1;
  Pointer second = 2;
}

message PerpendicularConstraint {
  Pointer first = 1;
  Pointer second = 2;
}

message MirrorConstraint {
  Pointer mirror = 1;
  message Pair {
    Pointer first = 1;
    Pointer second = 2;
  }
  repeated Pair mirrored_pairs = 2
      [(field_options).at_least = 1];
}

message DistanceConstraint {
  Pointer first = 1;
  Pointer second = 2;
  enum Direction {
    HORIZONTAL = 0;
    VERTICAL = 1;
    MINIMUM = 2;
  }
  Direction direction = 3;
  double length = 4;
  enum Alignment {
    ALIGNED = 0;
    ANTI_ALIGNED = 1;
  }
  enum HalfSpace {
    NOT_AVAILABLE = 0;
    LEFT = 1;
    RIGHT = 2;
  }
  message HalfSpaceParams {
    HalfSpace half_space_first = 1;
    HalfSpace half_space_second = 2;
  }


  oneof additional_params {
    option (oneof_options).handler =
        "select_distance_params";
    // The actual definition of the
    // handler is shown further in
    // the text.

    Alignment alignment = 5;
    HalfSpaceParams
        half_space_params = 6;
  }
}

message LengthConstraint {
  Pointer entity = 1;
  double length = 2;
}

message DiameterConstraint {
  Pointer entity = 1;
  double length = 2;
}

message RadiusConstraint {
  Pointer entity = 1;
  double length = 2;
}

message AngleConstraint {
  Pointer first = 1;
  Pointer second = 2;
  double angle = 3;
}

message HorizontalConstraint {
  repeated Pointer entities = 1
      [(field_options).at_least = 1];
}

message VerticalConstraint {
  repeated Pointer entities = 2
      [(field_options).at_least = 1];
}

message MidpointConstraint {
  Pointer midpoint = 1;
  message Endpoints {
    Pointer first = 1;
    Pointer second = 2;
  }
  oneof additional_params {
    Endpoints endpoints = 2;
    Pointer entity = 3;
  }
}
\end{lstlisting}

\begin{table*}[t!]
    \renewcommand{\arraystretch}{1.5}
    \centering
    \begin{tabular}{>{\ttfamily}r|l}
        \hline
        \textnormal{\textbf{Regular expression}} & \textbf{Range} \\
        \hline
        \multicolumn{2}{l}{Discrete tokens} \\
        \hline
        objects\textbackslash.kind & $ \{0, 1\} $ \\
        .*\textbackslash.entity.kind & $ \{0, 1, 2, 3\} $ \\
        .*\textbackslash.constraint.kind & $ \{0, \ldots, 15\} $ \\
        .*\textbackslash.is\_construction & $ \{0, 1\} $ \\
        .*\textbackslash.is\_clockwise & $ \{0, 1\} $ \\
        .*\textbackslash.is\_periodic & $ \{0, 1\} $ \\
        .*\textbackslash.additional\_params & $ \{0, 1\} $ \\
        .*\textbackslash.(entity|entities|first|second|midpoint|mirror) & $ \{0, \ldots, 255\} $ \\
        .*\textbackslash.direction & $ \{0, 1, 2\} $ \\
        .*\textbackslash.alignment & $ \{0, 1\} $ \\
        .*\textbackslash.half\_space\_(first|second) & $ \{0, 1, 2\} $ \\
        \hline
        \multicolumn{2}{l}{Continuous tokens} \\
        \hline
        .*\textbackslash.(point|start|end|center|interpolation\_points)\textbackslash.(x|y) & $ [-1.0, 1.0] $ \\
        .*\textbackslash.(start|end)\_derivative\textbackslash.(x|y) & $ [-100.0, 100.0] $ \\
        .*\textbackslash.(start|end)\_phi & $ [0.0, 3.0] $ \\
        .*\textbackslash.radius & $ [0.0, 1.0] $ \\
        .*\textbackslash.length & $ [0.0, 2 \sqrt{2}] $ \\
        .*\textbackslash.angle & $ [-10.0, 10.0] $ \\
    \end{tabular}
    \caption{\textbf{Token groups} used in our experiments. The \emph{left column} contains \texttt{python}-style regular expressions isolating specific subsets of tokens (\ie, dotted paths of the corresponding fields). The \emph{right column} shows the ranges of values for each group.}
    \label{tab:token_groups}
\end{table*}

\pagebreak

The handler in \texttt{DistanceConstraint} is defined as a \texttt{python} function:
\begin{lstlisting}[language=python,style=mycodestyle]
def select_distance_params(obj):
  if obj.direction in {"HORIZONTAL",
                       "VERTICAL"}:
    return 0
  else:
    return 1
\end{lstlisting}

\section{Token groups}
\label{sec:appendix_token_groups}

Table~\ref{tab:token_groups} details the specific grouping of tokens used in our experiments. Groups are defined by regular expressions matching subsets of dotted field identifiers. For each group we provide the range of possible values.

\section{Additional statistics}

\begin{figure*}[p]
    \centering
    \input{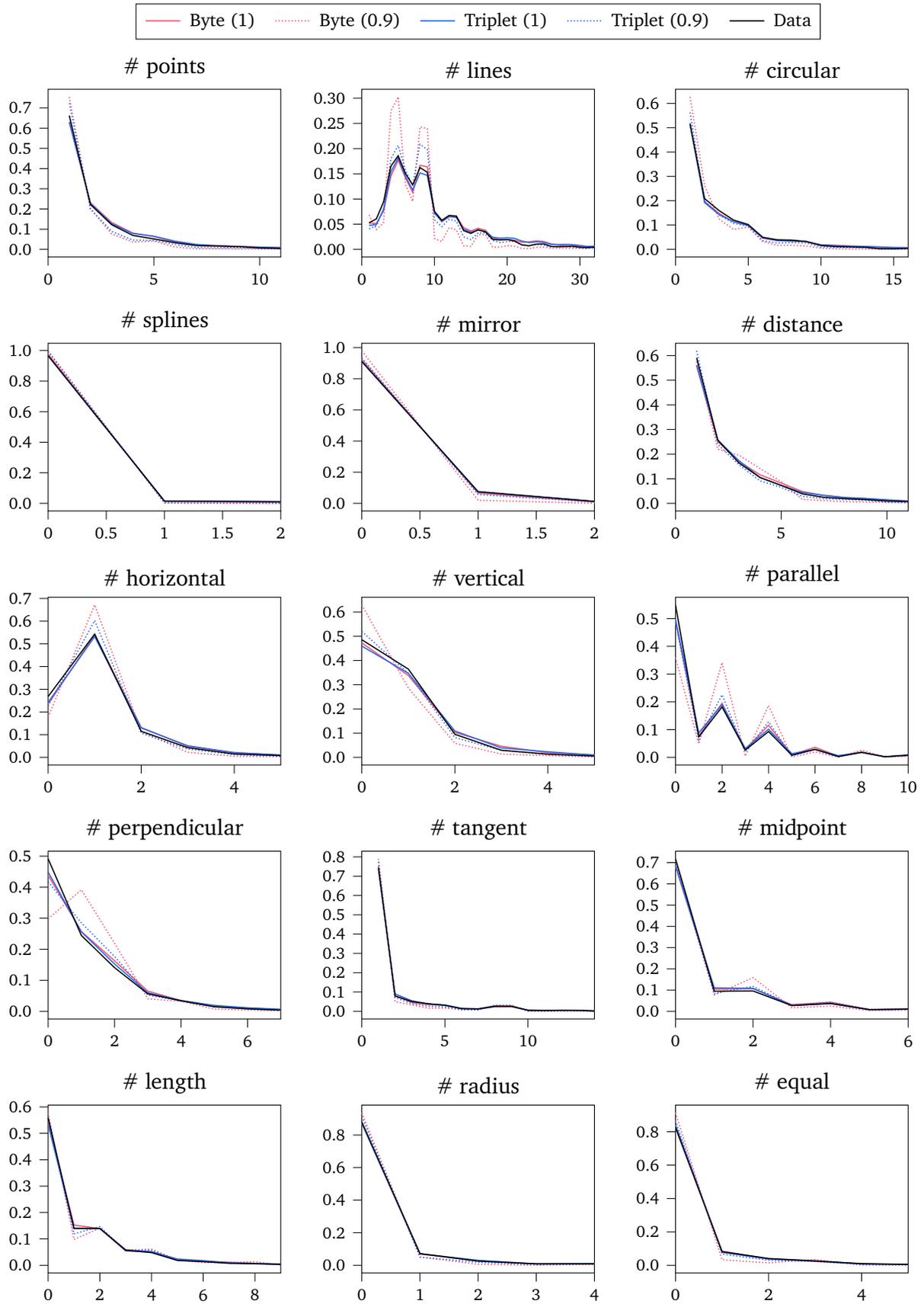}
    \caption{\textbf{Distribution of additional sketch statistics.} See also Figure~\ref{fig:stats_main}.}
    \label{fig:stats_appendix}
\end{figure*}

We present the distributions of a variety of additional sketch statistics in Figure~\ref{fig:stats_appendix}. These results are discussed in Section~\ref{sec:uncond_generation}.  

\end{document}